\documentclass{article}

\usepackage{microtype}
\usepackage{graphicx}
\usepackage{subfigure}
\usepackage{booktabs} % for professional tables

\usepackage{hyperref}

\usepackage[accepted]{icml2023}

\usepackage{amsmath}
\usepackage{amssymb}
\usepackage{mathtools}
\usepackage{amsthm}

\usepackage[capitalize,noabbrev]{cleveref}

\theoremstyle{plain}

\theoremstyle{definition}

\theoremstyle{remark}

\usepackage{amsmath}
\usepackage{amssymb}
\usepackage{mathtools}
\usepackage{amsthm}

\usepackage{graphicx}
\usepackage{multirow}
\usepackage{xcolor}

\definecolor{mycolor}{RGB}{198,223,218}

\usepackage{amssymb}% http://ctan.org/pkg/amssymb
\usepackage{pifont}% http://ctan.org/pkg/pifont
\newcommand{\cmark}{\ding{51}}%
\newcommand{\xmark}{\ding{55}}%
\usepackage{makecell}

\DeclareMathOperator*{\argmax}{arg\,max}
\DeclareMathOperator*{\argmin}{arg\,min}
\DeclareMathOperator*{\kl}{KL}
\DeclareMathOperator*{\MLE}{MLE}
\DeclareMathOperator*{\MSE}{MSE}
\DeclareMathOperator*{\vae}{VAE}
\usepackage[textsize=tiny]{todonotes}

\icmltitlerunning{Swapped goal-conditioned offline reinforcement learning}

\begin{document}

\twocolumn[
%\icmltitle{Generalize skills for offline goal-conditioned reinforcement learning}
\icmltitle{Swapped goal-conditioned offline reinforcement learning}

% It is OKAY to include author information, even for blind
% submissions: the style file will automatically remove it for you
% unless you've provided the [accepted] option to the icml2023
% package.

% List of affiliations: The first argument should be a (short)
% identifier you will use later to specify author affiliations
% Academic affiliations should list Department, University, City, Region, Country
% Industry affiliations should list Company, City, Region, Country

% You can specify symbols, otherwise they are numbered in order.
% Ideally, you should not use this facility. Affiliations will be numbered
% in order of appearance and this is the preferred way.
\icmlsetsymbol{equal}{*}

\begin{icmlauthorlist}
\icmlauthor{Wenyan Yang}{equal,tuni}
\icmlauthor{Huiling Wang}{tuni}
\icmlauthor{Dingding Cai}{tuni}
\icmlauthor{Joni Pajarinen}{aalto}
\icmlauthor{ Joni-Kristian K\"am\"ar\"ainen$^1$}{tuni}
%\icmlauthor{Firstname5 Lastname5}{yyy}
%\icmlauthor{Firstname6 Lastname6}{sch,yyy,comp}
%\icmlauthor{Firstname7 Lastname7}{comp}
%\icmlauthor{}{sch}
%\icmlauthor{Firstname8 Lastname8}{sch}
%\icmlauthor{Firstname8 Lastname8}{yyy,comp}
%\icmlauthor{}{sch}
%\icmlauthor{}{sch}
\end{icmlauthorlist}

\icmlaffiliation{tuni}{Computing Sciences, Tampere University, Finland}
\icmlaffiliation{aalto}{Department of Electrical Engineering and Automation, Aalto University, Finland}

\icmlcorrespondingauthor{Firstname1 Lastname1}{first1.last1@tuni.fi}
\icmlcorrespondingauthor{Firstname2 Lastname2}{first2.last2@aalto.fi}

% You may provide any keywords that you
% find helpful for describing your paper; these are used to populate
% the "keywords" metadata in the PDF but will not be shown in the document
\icmlkeywords{Machine Learning, ICML}

\vskip 0.3in
]

% this must go after the closing bracket ] following \twocolumn[ ...

% This command actually creates the footnote in the first column
% listing the affiliations and the copyright notice.
% The command takes one argument, which is text to display at the start of the footnote.
% The \icmlEqualContribution command is standard text for equal contribution.
% Remove it (just {}) if you do not need this facility.

\printAffiliationsAndNotice{}  % leave blank if no need to mention equal contribution
%\printAffiliationsAndNotice{\icmlEqualContribution} % otherwise use the standard text.

\begin{abstract}
Offline goal-conditioned reinforcement learning (GCRL) can be challenging due to overfitting to the given dataset.
% THESE ARE DODGY /Joni
%, resulting in extrapolation errors and an inability to learn generalizable policies.
To generalize agent's skills outside the given dataset, we propose a goal-swapping procedure that generates additional trajectories. To alleviate 
% WHAT MEANS LOW QUALITY TRAJECTORY UNCLEAR?
%the problem of low quality trajectories in the 
the problem of noise and extrapolation errors, we present
a general offline reinforcement learning  method called deterministic Q-advantage policy gradient (DQAPG).
% THIS IS CLEAR AS IT IS ALREADY SAID
%optimization that can learn a policy from noisy augmented data.
In the experiments, DQAPG outperforms state-of-the-art goal-conditioned offline RL methods in a wide range of benchmark tasks, and goal-swapping further improves the test results. It is noteworthy, that the proposed method obtains good performance on the challenging dexterous in-hand manipulation tasks for which the prior
methods failed.
%\textcolor{red}{address augmentation between successful trajectories}

%\textbf{Connect the words, "alleviate the problem"}
%Offline goal-conditioned reinforcement learning can be challenging due to the extrapolation error,  generalizability issue and sparse reward settings. In this work, we propose a simple yet data-efficient pipeline for offline goal-conditioned RL tasks, which contains two steps: goal-swapping augmentation and deterministic Q-advantage policy gradient (DQAPG) optimization. The goal-swapping is a general data-augmentation technique that handles the generalizability issue, and DQAPG can effectively learn policy from noisy augmented data. We validated our method's effectiveness through extensive tasks and outperformed prior state-of-the-art. Notably,  our method achieved the best performance on the most challenging Hand-manipulation tasks, while several prior arts failed.

%GoFAR’s training objectives can be re-purposed to learn an agent-independent goal-conditioned planner from purely offline source-domain data, which enables zero-shot transfer to new target domains. Through extensive experiments, we validate GoFAR’s effectiveness in various problem settings and tasks, significantly outperforming prior state-of-art. Notably, on a real robotic dexterous manipulation task, while no other method makes meaningful progress, GoFAR acquires complex manipulation behavior that successfully accomplishes diverse goals.

\end{abstract}

\section{Introduction}
\label{introduction}

Reinforcement learning (RL) has achieved remarkable success in a wide range of tasks, such as  game playing~\cite{mnih2013playing,game3,game4} and in robotics~\cite{robot1,robot2}. 
Goal-conditioned RL (GCRL)~\cite{GCRL,gcrl2,HER} allows us to learn a more general RL agent which can reach an arbitrary goal without retraining.
%researchers extend the classical RL problem towards multi-task learning problems, also known as goal-conditioned reinforcement learning (GCRL)~\cite{GCRL, gcrl2,HER}. 
Although general policy learning is appealing, training goal-conditioned RL can be difficult. GCRL tasks usually have sparse rewards, and therefore the agent needs to explore the environment intensively, which is infeasible and dangerous in many real-world applications. On the other hand, offline RL has become a research hotspot in recent years as it learns the policy from offline datasets without endangering the real environment~\cite{offline1,prudencio2022survey}. %Offline RL does not need to interact with the environment but purely learns from an offline dataset.
The combination of offline RL and goal-conditioned RL takes the best of both worlds,
\begin{figure}[h]
    \centering
    \includegraphics[width=.45\textwidth]{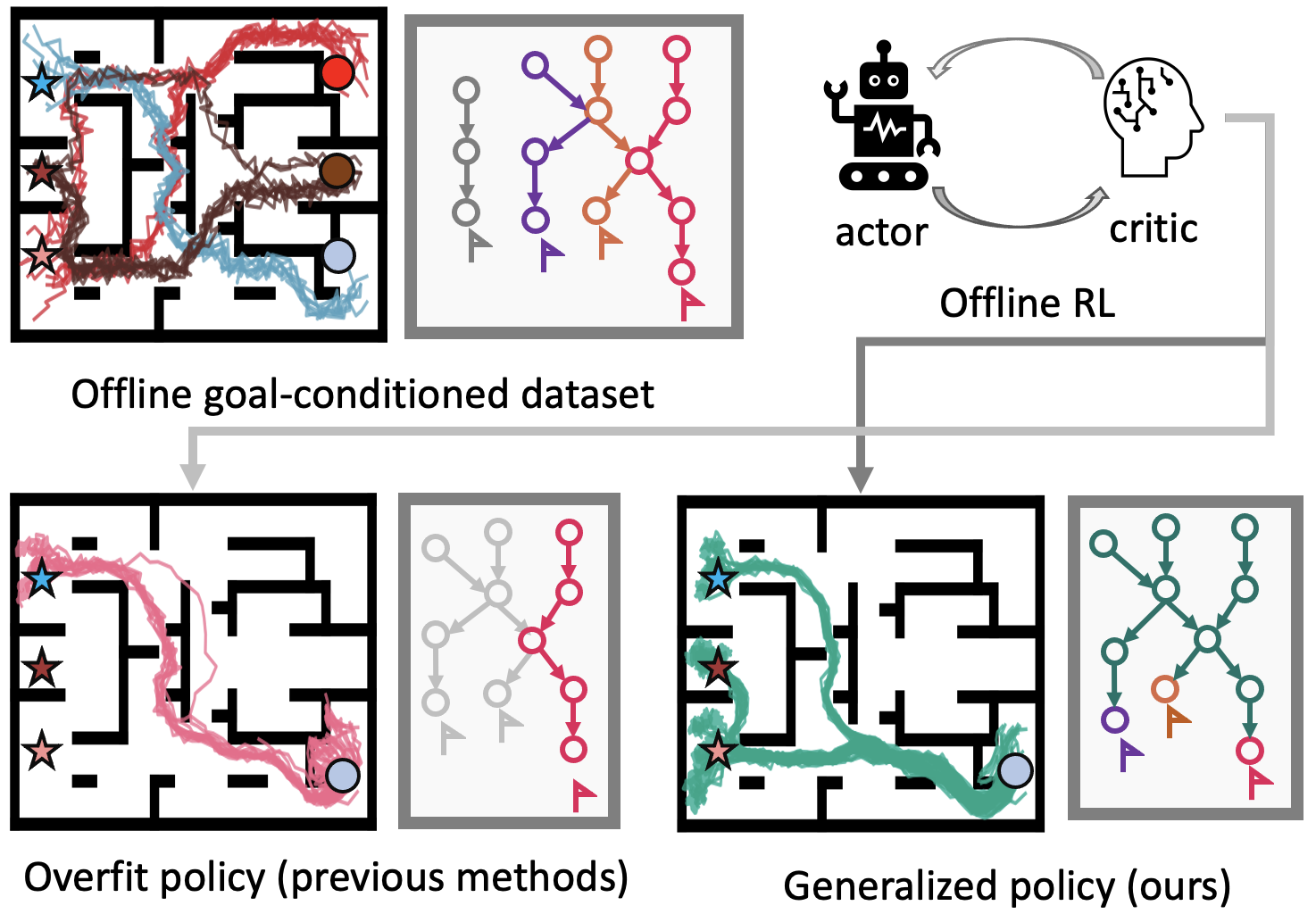}
    \caption{
    Illustration of the learned policy of our (below green) and previous (below red) offline GCRL learning. Our method learns a more general policy that reaches the goals from multiple states while the previous ones overfit to the training trajectories (top).}
    \label{fig:intro}
\vspace{-10pt}
\end{figure}
generalization and data efficiency, and is a promising approach for real-world applications~\cite{GoFAR}.

%However, while offline goal-conditioned RL (GCRL) combines the advantages of both offline RL and goal-conditioned settings it also inherits the challenges.
The offline goal-conditioned RL (offline GCRL) meets two new challenges. At first, during training,
the policy is likely to generate actions that are not present in the offline dataset.
%From the offline RL perspective, the distributional issue is inevitable as the agent cannot explore the environment. 
%During training, the policy is likely to generate actions that are not contained in the dataset. 
The value function cannot be correctly estimated for out-of-distribution actions. The biased value function causes the policy to deviate due to compounding errors~\cite{offline1}. Prior works apply policy or value function constraints to solve the out-of-distribution problem.
% LETS NOT RENAME THE PROBLEM
%is distributional shift issue.
For example, the prior approaches constrain the policy to generate actions within the dataset or make value function updates conservatively. The constraints unfortunately limit the policy's performance~\cite{offline1,prudencio2022survey}. 
The second challenge of offline GCRL is that each state can be assigned multiple goals, which means that the number of possible state-goal combinations can be extensive. Now general policy learning
becomes difficult as the dataset covers only a limited state-goal observation space~\cite{AM}.
 To solve the offline GCRL learning problem, prior works apply hindsight labeling to generate positive goal-conditioned observations within all sub-sequences~\cite{GCSL,WGCSL,GoFAR,HER}. However, hindsight relabelling has only limited positive impact on general skill learning, and the methods often learn an over-fitted goal-conditioned policy as shown in Figure~\ref{fig:intro}. \citet{AM} propose a goal-chaining technique to learn skills across trajectories, but it cannot handle the resulting noisy data properly~\cite{GoFAR}. 

In this work, we propose simple and efficient solutions to the challenges of offline GCRL: 1) goal-swapping, and 2) deterministic Q-Advantage policy gradient (DQAPG). Goal-swapping is a data augmentation technique that expands the dataset and thus provides a larger offline space to sample. DQAPG is an adaptive policy-constrained offline RL method that effectively learns from noisy samples. To summarize, our contributions are:
\begin{itemize}
    \item   We formulate a goal-swapping data augmentation technique, which enables an agent to learn more general skills through the given offline dataset trajectories.
\item  We propose an adaptive policy constraint offline RL  method that learns a policy more effectively from noisy trajectories. %Our proposed DQAPG is an adaptive policy constraint-based offline RL solution that applies less conservative policy constraints than prior similar solutions. 
\end{itemize}
The proposed methods are evaluated on
a wide set of offline GCRL benchmark tasks.
The methods achieve superior performance compared to the available baselines, especially in the most challenging dexterous in-hand manipulation tasks.

\section{Preliminaries}
\label{subsec:preliminary}

\textbf{Goal-conditioned Markov decision process.\quad}
The classical Markov decision process (MDP) $\mathcal{M}$ is defined as a tuple  $<\mathcal{S}, \mathcal{A}, \mathcal{T}, r, \gamma, \rho_0>$, where  $\mathcal{S}$ and $\mathcal{A}$ denote state space and action space, $\rho_0$ represents the initial states' distribution,  $r$ is the reward, $\gamma$ is the discount factor, and $\mathcal{T}$ denotes the state transition function ~\cite{RL}. %The objective is to learn a policy $\pi(a|s):\mathcal{S} \rightarrow \mathcal{A}$ that maximize the expected cumulative return:
%\begin{equation}
%\mathcal{J}(\pi) = \mathbb{E}_{a_t\sim\pi(\cdot|s_t),s\sim \rho_0, s_{t+1}\sim \mathcal{T}(\cdot|s_t, a_t)} [\sum_{t=0}^{\infty}\gamma^t r(s_t,a_t)]
%\end{equation}
For goal-conditioned tasks, one additional vector $g$ specifies the goal that the agent should achieve. 
%The goal can be considered a description of the final state.  The representation of the goal does not have unique forms, as long as it describes the property of the objective, such as image-based goals~\cite{}, language-based descriptions ~\cite{}, etc. 
In general, the goal-conditioned RL augments MDP with the extra tuple $<\mathcal{G}, p_g, \phi>$, where $\mathcal{G}$ is the goal space and $p_g$ is the distribution of the desired task goals. $\phi: \mathcal{S}\rightarrow\mathcal{G}$  is a tractable mapping function that maps the state to a specific goal.
%\begin{equation}
%label{eq:goal-func}
%    \phi:\mathcal{S} \rightarrow \mathcal{G}, s\in\mathcal{G}, s\in\mathcal{S},
%\end{equation}
 The state-goal pair $(s,g)$ forms a new observation and is used as the input for agent $\pi(a|s,g)$. 
%In some scenarios (e.g., PointMaze, etc.), the goal is identical to state. 
The goal-conditioned MDP (GC-MDP) is represented as $<\mathcal{S}, \mathcal{G}, \mathcal{A}, \mathcal{T}, r, \gamma, \rho_0, p_g >$~\cite{GCRL} (shown in Fig.\ref{fig:goal-conditioned MDP}).  
%More specifically, the goal-conditioned reward is a binary reward  \[r_t = R(\phi(s),g)=R(ag_t,g)=\mathbf{1}[ag_t=g],g\in\mathcal{G},\] 
%where $\mathbf{1}$ is the indicator function that checks if goal is reached.
The policy $\pi(a|s,g)$ makes decisions based on the state-goal pairs. The objective of GC-MDP can be formulated as:
\begin{figure}[h]
    \centering
    \includegraphics[width=.49\textwidth]{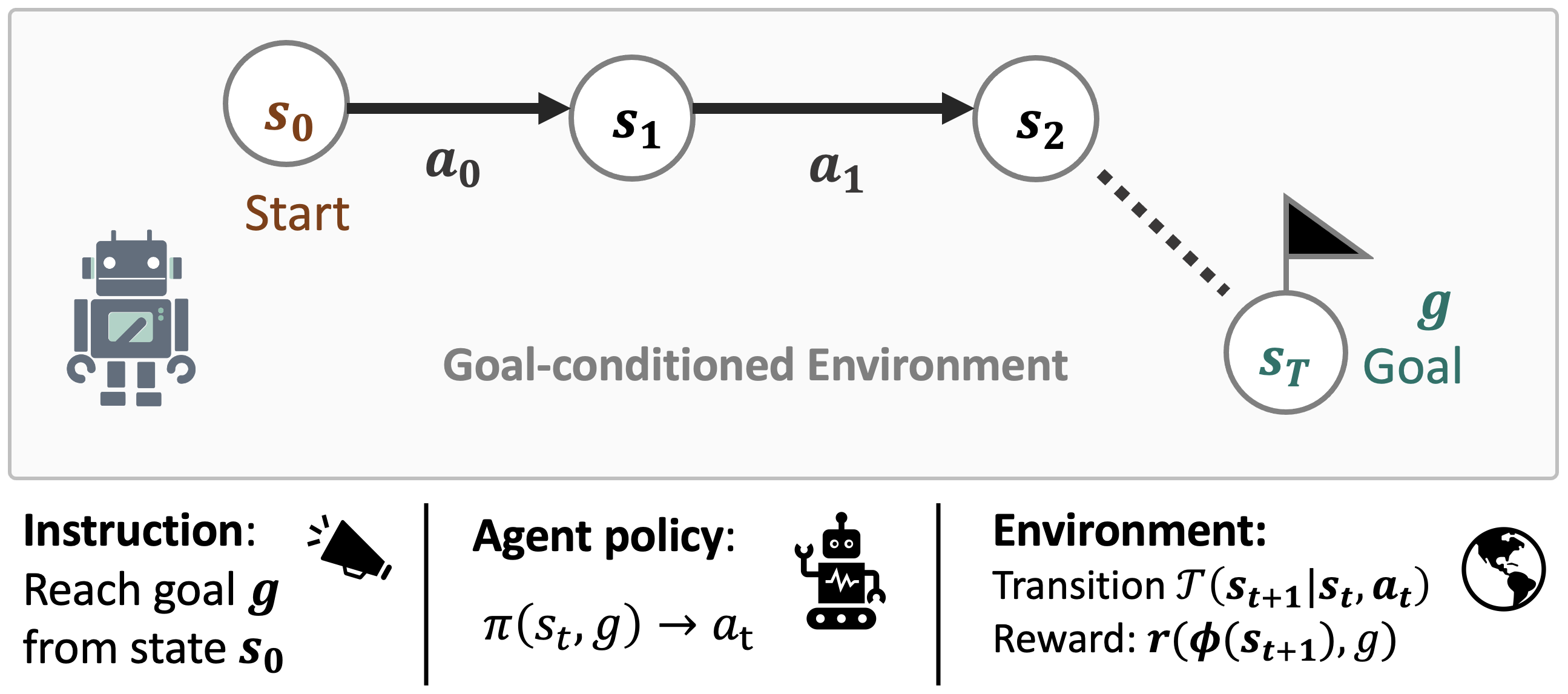}
    \caption{Visualization of the goal-conditioned Markov decision process (GC-MDP). Each state $s$ has a corresponding representation in the goal space $\eta=\phi(s)$.  }
    % Each state $s$ has a corresponding goal $\eta=\phi(s)$.
    \label{fig:goal-conditioned MDP}
\end{figure}
\begin{equation}
\label{eq:goal-conditioned MDP-objetcive}
\mathcal{J}(\pi) = \underset{a_t\sim\pi(\cdot|s_t,g),g\sim p_g, s_{t+1}\sim \mathcal{T}(\cdot|s_t, a_t)}{\mathbb{E}} \left[\sum_{t=0}^{\infty}\gamma^t r_t\right].
\end{equation}
Two value functions are defined to represent the expected cumulative return, state-action value $Q$, and state value $V$. For GC-MDP, we have  $V^\pi(s,g)$ function
%\[V^\pi(s,g)=\mathbb{E}_{a_t\sim\pi(\cdot|s_t,g)}[\sum_{t=0}^{\infty}\gamma^t r_t | \mathcal{S}_t=s,g],\] 
that is the goal-conditioned expected total discounted return from observation pair $(s,g)$ using policy $\pi$,  and $Q^\pi(s_t,a_t,g)$ function
%\[Q^\pi(s_t,a_t,g)=\mathbb{E}_{a_t\sim\pi(\cdot|s_t,g)}[\sum_{t=0}^{\infty}\gamma^t r_t | \mathcal{S}_t=s,\mathcal{A}_t=a,g]\]
 estimates the expected return of an observation $(s_t,g)$ for the
 action $a_t$ for the policy $\pi$.  Furthermore, we have an advantage function $A^\pi(s,g,a) = Q^\pi(s,g,a) - V^\pi(s,g)$,
which is another version of the Q-value but with lower variance. When the optimal policy $\pi^*$ is obtained, $Q^*(s,g,a)=V^*(s,g)$~\cite{RL}.  In this work, we define a sparse reward
\begin{equation}
\label{eq:r}
        r(s,g)= 
\begin{cases}
    0, & \text{if } ||\phi(s), g|| < \epsilon\\
    -1,              & \text{otherwise}
\end{cases},
\end{equation} 
where $||\phi(s), g||$ is a distance metric measurement, and $\epsilon$ is a distance threshold. We set the discount factor $\gamma=1$. 
In such case, $V(s,g)$ represents the expected horizon from state $s$ to goal $g$, and $Q(s,g,a)$ represents the expected horizon from state $s$ to the goal $g$ if an action $a$ is taken. This setting produces an intuitive objective:  \textit{Find the policy that takes the minimum number of steps to achieve the task's goal.}

%\paragraph{Actor-critic policy gradient}
%Given a policy $\pi_\psi(s,a)$ with parameters $\psi$, the objective to find  $\psi$ that maximizes the expected total discounted return $\mathcal{J}(\pi)$. For continuous control tasks, we can simply optimize the parameter $\psi$  in the direction of the performance gradient $\nabla_\psi \mathcal{J}(\pi_\psi)$, which is called policy gradient theorem~\cite{sutton}. For a stochastic policy in goal-conditioned settings, there are several equivalent forms of policy gradient ~\cite{pg}:
%\begin{align*}
%    \nabla_\psi \mathcal{J}(\pi_\psi) = \mathbb{E}_{a~\sim\pi_\psi} [\nabla_\psi \log \pi_\psi(a|s,g) V(s,g)]\\ %\Upsilon
%    =  \mathbb{E}_{a~\sim\pi_\psi} [\nabla_\psi \log \pi_\psi(a|s,g) Q(s,g,a)] \\
%    = \mathbb{E}_{a~\sim\pi_\psi} [\nabla_\psi \log \pi_\psi(a|s,g) A(s,g,a)].
%\end{align*}
%The value function plays as a critic that weights the likelihood ratio of policy's actions leads to better performance. Such policy gradient method is also called actor-critic policy gradient. Alternatively, we can directly use the gradient of value function $\Q$ to optimize the policy $\pi$ if the policy is  detetministic : $\pi(s,g)\rightarrow a$. In such case, the deterministic policy gradient in goal-conditioned RL can be written as:
%\begin{equation*}
%    \nabla_\psi \mathcal{J}(\pi_\psi) = \mathbb{E}_{s\sim \mathcal{T},\pi_\psi}[\nabla_\psi\pi_\psi(s)\nabla_aQ_\theta(s,a,g)|_{a=\pi_\psi(s)}]
%\end{equation*}

\textbf{Offline goal-conditioned reinforcement learning.\quad}
Offline reinforcement learning (offline RL) is a specific problem of reinforcement learning where the agent cannot interact with the environment but only access a static offline dataset collected by unknown policies~\cite{offline1,prudencio2022survey}.  The objective of offline goal-conditioned RL setting is the same as online goal-conditioned RL defined in Equation~\ref{eq:goal-conditioned MDP-objetcive} but without interacting with the environment.
Based on the definition of GC-MDP, we formulate the offline goal-conditioned RL dataset as  $\mathcal{D}:=\{\mathbf{\zeta}_i\}_{i=1}^N$, where $\mathbf{\zeta}$ is the goal-conditioned trajectory and $N$ is the number of stored trajectories  
\begin{align*}
    \mathbf{\zeta}_i = \{<s_0^i, \eta_0^i, a_0^i, r_0^i>, <s_1^i, \eta_1^i, a_1^i, r_1^i>,\\
    ..., <s_T^i, \eta_T^i, a_T^i, r_T^i>, g^i\} \enspace .
\end{align*}
$\eta$ is the state's corresponding goal representation calculated using $\eta_t = \phi(s_t)$. The task goal $g^i$ is randomly sampled from $p_g$ and the initial state $s_0\sim\rho_0$. Note that the trajectories can be unsuccessful trajectories ($\eta_T^i \neq g^i$).

%\paragraph{Expectile regression.} We first introduce expectile regressions, 

%Practical methods for estimating various statistics of a random variable have been thoroughly studies in applied statistics and econometrics. The $\tau \in [0,1]$ expectile of some random variable X is defined as a solution to the asymmetric least squares problem:
%\[\underset{m_\tau}{\argmin} \mathbb{E}_{x\sim X} of a conditional distribution:
%\[\underset{m_\tau(x)}{\argmin} \mathbb{E}_{(x,y)\sim D} [L_2^\tau(y-m_\tau(x))]\]

\section{Challenges and related works}

In this section, we discuss the challenges in the goal-conditioned offline RL setting. We first discuss the generalization issue and the importance of data utilization. About the offline RL aspect, we discuss the extrapolation problem and analyze the pros and cons of prior solutions.

\subsection{Generalization power of goal-conditioned RL}
\label{subsec:gcrl_challenge}
%In goal-conditioned RL,  different goals $g\sim p_g$ implies different goal-conditioned tasks under the same environment. The agent must learn a unified policy to perform under multiple goals simultaneously. 
Essentially, the objective of GCRL is to learn a general policy that can achieve all reachable goals~\cite{GCRL}. The agent must learn a unified policy to perform multiple goals. However, the offline dataset state-goal pairs only cover a limited space of the goal-conditioned MDP. In other words,  if we train a policy with the offline dataset,  the policy learns to reach goals within a single trajectory in the dataset. This solution is not general, but undesirably specific.
We present a more detailed discussion in Section~\ref{subsec:goal-swap-augmentation}. To learn a generalized goal-conditioned policy, the work actionable models (AM)~\cite{AM} propose a goal-chaining technique that conducts goal-swapping augmentation. It assigns conservative values to the augmented out-of-distribution data for Q-learning. However, AM's performance is limited when the dataset contains noisy data labels ~\cite{GoFAR}. %Some works also proposed similar techniques but are limited to online settings~\cite{check AM references}. 
Besides goal-chaining, hindsight relabelling, or hindsight experience replay (HER), is also helpful in improving the sample efficiency for online goal-conditioned RLs~\cite{HER,WGCSL}. It relabels trajectory goals to states that were actually achieved instead of the task-commanded goals. Although HER efficiently utilizes the data within a single trajectory, it cannot connect different trajectories as goal-chaining does.

\subsection{Offline reinforcement learning challenges} 
 \label{subsec:offline_rl_related}

As arbitrary policies collect the offline dataset, applying off-policy methods for offline RL problems is intuitive. 
However, as the agent cannot access the environment, the off-policy RL usually faces distributional shift problems. The policy $\pi_\psi$ will likely generate actions that are not contained in the offline dataset. 
%and the Q-function will be trained under the out-of-distribution state-action pairs. 
It makes Q-function unable to estimate the values correctly, resulting in bigger mistakes that compound once the policy diverges wildly from the dataset. This distributional shift issue is also called extrapolation error~\cite{offline1, prudencio2022survey}. Generally, researchers mainly propose three types of methods to solve offline RL problems~\cite{prudencio2022survey}:  policy constraint methods, value function constraint methods, and one-step RLs. Policy constraint methods force the learned policy $\pi_\psi$ to stay close to the offline dataset's behavioral policy $\pi_D$, either by explicitly sampling actions from $D$~\cite{BCQ,wu2019behavior} or minimizing certain divergence measurements between $\pi_\psi(\cdot|s)$ and $\pi_D(\cdot|s)$~\cite{BEAR, TD3BC, RWR, AWR,AWAC,CRR,kostrikov2021offline}. Value constraint methods add regularization terms on the value function $Q$ to have a more conservative value estimation. It pushes up state-action pairs' values in the dataset and pulls down the values in unseen actions~\cite{CQL,fujimoto2019benchmarking}. 
%Such value function constraint approaches tend to be less conservative on policy compared with policy constraint methods. 
In some challenging tasks (e.g., Kitchen tasks~\cite{D4RL}), the value constraint approaches usually outperform policy constraint methods because the value function constraint approaches tend to be less conservative on policy than policy constraint methods. Some works also propose to use one-step on-policy RL  to avoid extrapolation error. These works only do policy-evaluation on the offline dataset instead of on the policy. However, these methods can hardly learn the optimal value function due to the on-policy nature~\cite{onestep, IQL,VEM}.

In summary,  the offline goal-conditioned RL not only faces extrapolation issues but also requires the policy to learn a generalizable policy that can reach all possible goals within the dataset. In this work, we aim to design a generalizable data-efficient offline RL solution for offline GCRL settings. 

%for policy, the policy will generate actions not contained in the dataset during the training, and it induces  generalization error in the approximate value function. %The generalization error also makes it difficult to evaluate the expected value of a policy which is sufficiently different from the behavior policy. 

%The main challenge of offline goal-conditioned RL is its nature of sparse reward. As is defined in Eq.\ref{eq:goal-func},  reward function is a simple unshaped binary signal that indicate if the goal is achieved. The unsuccessful trajectories will not be assigned with any reward signal, which makes the agent hardly learn any useful epxerience to achieve the goal.

%Based on  the definition of goal-conditioned MDP (sec.\ref{subsec:preliminary}), the offline goal-conditioned RL potentially is more likely to have generalizability issue than offline RL. As the dimension inputs of increased (from state only to state-goal pairs)

%\paragraph{Sparse reward} As is defined in Eq.\ref{eq:goal-func},  reward function is a simple unshaped binary signal that indicate if the goal is achieved. Practially, the reward function is defined to compare the goal distances to assign reward. The unsuccessful trajectories will not be assigned with any reward signal, which makes the agent hardly learn any useful epxerience to achieve the goal.

\section{Method}

\begin{figure*}[h]
    \centering
    \includegraphics[width=1.\textwidth]{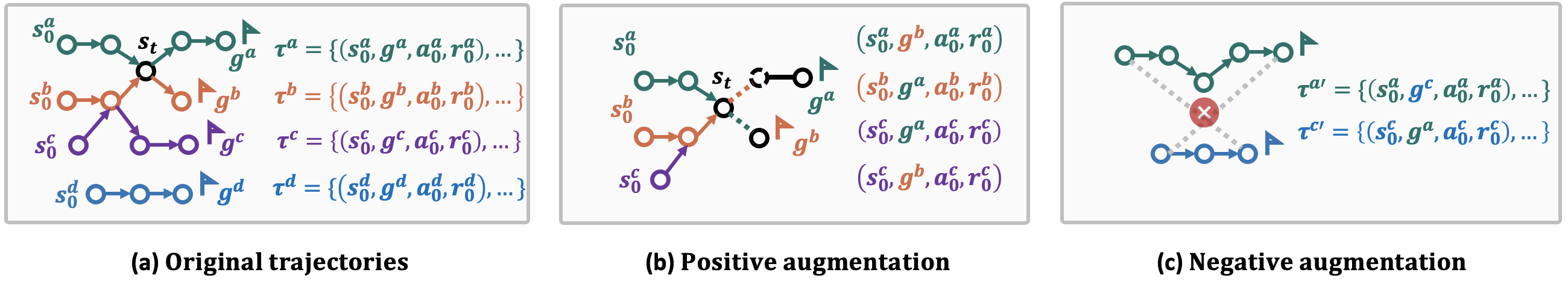}
    \caption{Example of the goal-swapping data augmentation. (a) original offline trajectories (denoted by four different colors). (b) augmentation example where the goals are reachable in each generated trajectory (positive augmentation). (c) negative augmentation where the goals are not anymore reachable.}%$\tau^a$ and $\tau^b$ have mutual visited state $s_t$. The
    \label{fig:goal-swap-analyze}
\end{figure*}

\begin{figure*}[h]
    \centering
    \includegraphics[width=1.\textwidth]{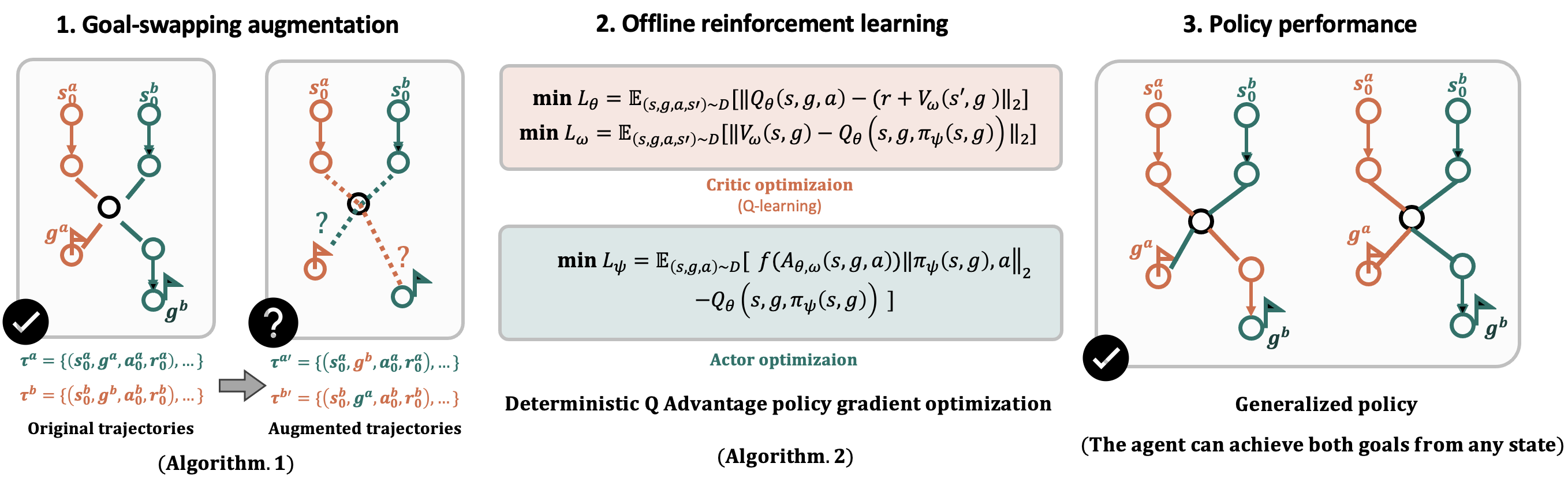}
    \caption{The schematic illustration of our framework. Firstly the goal-swapping data augmentation increases the amount of offline training pairs (Algorithms.~\ref{alg:goal-swap}). Then DQAPG then learns a general policy that can achieve different goals from the noisy training pairs (Algorithms.~\ref{alg:DQAPG}).  Finally, the dynamic programming  nature of Q-learning learns a policy to achieve all achievable goals within the dataset.}
    \label{fig:overview}
\end{figure*}

 To solve the challenges discussed in the previous section, we propose two simple yet effective methods:
 1) \textit{goal-swapping augmentation} and
 2) \textit{deterministic Q-Advantage policy gradient} (DQAPG). A schematic illustration of the methods inside the RL framework is  Figure~\ref{fig:overview}. The agent first conducts goal-swapping data augmentation to generate new state-goal  pairs from the existing trajectories. Then  DQAPG  optimizes the policy from the augmented noisy data pairs. 
 %Our proposed DQAPG is an adaptive policy constraint-based offline RL solution that applies less conservative policy constraints than prior similar solutions. 
 For this to work, we make an assumption that in the same environment, all goals are reachable from all states.
% The proposed methods can effectively learn a generalized policy. This section will illustrate the details of our  methods.
  
 %First, goal-swapping augmentation can improve the policy's generalizability by doing data augmentation in state-goal space. Second, DQAPG filters the non-optimal noisy augmented data pairs, meanwhile handles the extrapolation error for offline policy optimization.  We provide a schematic illustration in Figure~\ref{fig:overview}: the agent first conducts goal-swapping data augmentation to generate new state-goal observation pairs. We then use DQAPG to optimize the policy from augmented noisy data pairs. Finally, the optimized approach shows better generalizability in goal-conditioned RL tasks and can handle certain disturbances. This section provide the details of our proposed method.

\subsection{Goal-swapping data augmentation}
\label{subsec:goal-swap-augmentation}

As discussed in Section~\ref{subsec:gcrl_challenge}, the offline goal-conditioned dataset contains only a limited number of state-goal observations. 
%As is defined in goal-conditioned MDP, the policy receives both state-goal pairs $\{s,g\}$ as inputs to generate goal-conditioned actions. In such case, if we 
Therefore, training with offline data often results in overfitted policies.
%which would only learn to reach goals reachable within a single trajectory $\tau$. 
An illustrative example is in Figure~\ref{fig:goal-swap-analyze}(a). In this example environment, the goals $g_a$ and $g_b$, are (forward) reachable from the three states $s_0^a$, $s_0^b$ and $s_0^c$. However, the original conditioned trajectories $\tau^a$ and $\tau^b$ limit the agent's exploration to the known state-goal pairs
$s_0^a\rightarrow g^a$,
$s_0^b\rightarrow g^b$, and
$s_0^c\rightarrow g^c$.
%, and
%can only achieve the goals $g^a$ and $g^b$ from the states $s_0^a$ and $s_0^b$, respectively.
Our goal-swapping (see Figure~\ref{fig:goal-swap-comparison}(b)) avoids
the problem in an intuitive and simple-to-implement way.
%It is worth noting that if we swap the goals between these two trajectories, the new generated goal-conditioned trajectories are 

To make the agent achieve as many goals as possible, we formulate a general goal-swapping augmented experience replay technique sketched in Algorithm~\ref{alg:goal-swap}.
The algorithm randomly samples two trajectories and then swaps the goals between these two. In this way, it generates up to a combinatory number of new goal-conditioned trajectories.  The dynamic programming nature of RL connects the state-goal pairs across different trajectories.

%Note that the data-agumented goal-conditioned trajectory 

 %We now illustrate  and discuss two cases of such augmentation,  positive augmentation and negative augmentation, and how they affect the training of offline goal-conditioned RL.

\textbf{Goal-swapping analysis.\quad} Let's continue the example in Figure~\ref{fig:goal-swap-analyze}(a),
where the goals $g\in[g^a,g^b, g^c]$ are reachable from the states $s\in [s_0^a, s_0^b, s_0^c]$ although not explicitly present in the offline dataset. However, if a goal is swapped between the original three trajectories $[\tau^a, \tau^b, \tau^c]$, the augmented goal-conditioned tuples shown in Figure~\ref{fig:goal-swap-analyze}(b) become available. The offline RL methods can chain the new trajectories thanks to the dynamic programming part of TD learning.  Let's
take one value estimation as an example, $V(s_t,g)=r+V(s_{t+1},g)$. For the pair $(s_{t+1},g)$, as long as $g$ is reachable and the value $V(s_{t+1},g)$ is well approximated, TD learning can backpropagate the state-goal (state-goal-action) values to the previous pairs. An illustration is in Fig.\ref{fig:goal-swap-analyze}(b), where the state $s_t$ is the "hub state" shared by the three original trajectories and from which all goals $g^i\sim[g^a, g^b, g^c]$ are reachable. The values of these goal-conditioned states, $Q(s_t, a_t^i, g^i)$, can be estimated and recursively backpropagated to $Q(s_0, a_0^i, g^i), i \in [a,b,c]$.  Goal-swapping creates as many state-goal pairs as possible, so that TD learning ultimately backpropagates values over all trajectory combinations.

Although offline RL TD learning can connect goals across trajectories, applying the goal-swapping augmentation technique can be tricky. The reason is that the goal-swapping process is random, creating many non-optimal state-action pairs. Furthermore, those augmented state-action pairs may not even have a solution (the augmented goals are not reachable) within the offline dataset.
This is demonstrated in Fig.~\ref{fig:goal-swap-analyze}(c).  
For this purpose, we rely on
the \textit{actionable model} (AM)~\cite{AM}, where
 Q-learning is heavily modified in a conservative Q-learning manner to handle the negative state-goal observations. This raises another problem as the performance of AM is very limited on noisy datasets~\cite{GoFAR}. 

%The noisy data issue is addressed next in Section~\ref{subsec:our_policy_gradient}.

\begin{algorithm}[tb]
    \caption{Goal-swapping augmented experience replay}\label{alg:goal-swap}
    \begin{algorithmic}[1]
       \STATE {\bfseries Requiure:} offline dataset as $D$, goal-conditioned tuples  $\zeta$,  state-goal mapping function $g_i=\phi(s_i)$, and reward function $r=R(\phi(s),g)$. 
    
       \STATE Sample goal-conditioned transitions $\zeta$ from $D$: \\ $\zeta_i = \{g,s,a,r,s'\} \sim D.$
       \STATE Sample random goals: $g_{rand} \sim D$
       \STATE Generate $\tau_{rand}$  by replacing $g$ with $g_{rand}$ in $\zeta$:
        \\$\zeta_{aug}=\{g_{rand},s,a,r_{aug},s'\}$
        \\where $r_{aug}=R(\phi(s'),g_{rand})$ 
        \STATE Return $\zeta$ and $\zeta_{aug}$ 
    \end{algorithmic}
\end{algorithm}
%To fully utilize the offline goal-conditioned data, we applied the goal-swapping augmentation together with hindsight relabeling for policy learning.

\subsection{Deterministic Q-Advantage policy gradient}\label{subsec:our_policy_gradient}

To effectively learn from augmented offline data that inherently produces noisy data samples, we propose a deterministic Q-Advantage policy gradient method (DQAPG). The method
is built upon the standard deterministic policy gradient~\cite{dpg}. For better illustration, we denote the parameter of $V(s,g)$ as $\phi$, the parameter of $Q(s,g,a)$ as $\theta$ and $\pi(s,g)$ is parameterized by $\psi$.

\textbf{Value estimation.\quad}
We first aim at learning the optimal value $V_\phi(s)$ function and Q-function $Q_\theta(s,a)$ from the offline dataset.
Based on the definition, the optimal $Q^*$ value equals to the optimal $V^*$ value (for the optimal policy $\pi^\ast$):
\[Q^*(s,a)=V^*(s)\enspace .\]
This indicates that $V(s,g)$ value function estimation can be added to the deterministic policy gradient (DPG) framework to learn the optimal $V^*$. Now we have the objective of the $V$ function learning: 
\begin{equation}
    \label{eq:v_adv}
    L(\omega)=\underset{(s,g,a)\sim D}{\mathbb{E}}[\lVert V_\omega(s,g) - Q_\theta(s,g,\pi_\psi(s,g)){\rVert}_2]
\end{equation}
and the corresponding objective of the $Q_\theta$ learning is
\begin{equation}
    \label{eq:q_adv}
    L(\theta)=\underset{(s,g,a,s')\sim D}{\mathbb{E}}[\lVert Q_\theta(s,g,a) - (r+V_\omega(s',g)){\rVert}_2] \enspace .
\end{equation}
Based on the unity reward in Eq. ~\ref{eq:r}, we set the optimization constraints
$-H\leq V \leq 0$ and $-H\leq Q \leq 0$, where $H$ represents the maximum task horizon.

\textbf{Policy optimization.\quad}
To suppress the extrapolation errors that can be dramatic in offline RL with generated data, we also constraint the policy similar to policy constraint offline RL methods. This is achieved by applying a regularization term that enforces
$\pi_\psi(s,g)$ to be close to the actions stored in the dataset. This is implemented as the following offline RL objective~\cite{AWAC, AWR}:
\begin{equation}
\label{eq:policy_constraint_obj}
  \begin{split}
    \pi_{k+1} = \underset{\pi \in \Pi}{\argmax} \underset{(s,g,s')\sim D, a\sim\pi(s,g)}{\mathbb{E}}\left[ \mathbb{Q}^{\pi_k}(s,g,a)\right] \\
   \hbox{s.t.}~\kl\left(\pi (\cdot|s,g)\lVert \pi_D(\cdot|s,g)\right)<\epsilon,
\end{split}  
\end{equation}
%Such objective has been widely used in works~\cite{AWAC, CRR, etc.}.
where $\kl$ is the KL-divergence and $\pi_D$ is a behavioral policy that is obtained using only the dataset samples $D$. $\mathbb{Q}$ can be the Q-function $Q(s,g,a)$ or the advantage function $A(s,g,a)$~\cite{prudencio2022survey,RL}. 
However, different choices can result in different objectives. 
First, if we use the advantage function $A(s,g,a)$ as $\mathbb{Q}$ in Eq.~\ref{eq:policy_constraint_obj}, we can derive the following objective for a deterministic policy:
\begin{multline}
\label{eq:adv_bc}
     L(\psi)_{awbc}= \underset{(s,g,a)\sim D}{\mathbb{E}}[ \exp(A^{\pi_\psi}(s,g,a)) \\ \cdot  \lVert \pi_\psi(s,g)-a{\rVert}_2]. 
\end{multline}
If we choose to optimize $Q(s,g,a)$ in Equation~\ref{eq:policy_constraint_obj}, we  have:
\begin{equation}
    \label{eq:td3bc_loss}
    L(\psi)_{qbc}=\underset{(s,g,a)\sim D}{\mathbb{E}}[ \lVert \pi_\psi(s,g)-a\rVert_2 - Q^{\pi_\psi}(s,g,\pi_\psi(s))].
\end{equation}
The details above can be found in Appendix~\ref{subsec:adv_obj_inference}.
The two objectives optimize the policy differently. The objective $L(\psi)_{awbc}$ is advantage-weighted behavioral cloning (BC), and is widely used in prior offline RL~\cite{AWAC,CRR,AWR}. However, such policy constraint tends to be more pessimistic as the actions in the dataset may not be the optimal goal-conditioned actions~\cite{prudencio2022survey}. The objective $L(\psi)_{qbc}$ optimizes the policy through deterministic policy gradient, similarly justified in~\cite{TD3BC}. 
%can be considered as an advantage weighted behavioral cloning, where the weights is adaptively adjusted by the advantage function.
%However, such policy constraint tend to be more pessimistic as the actions in the dataset  may not be the optimal goal-conditioned actions~\cite{prudencio2022survey}. Furthermore, goal-swapping augmentations  are likely to be non-optimal and simply using $L(\psi)_{awbc}$ may not be beneficial.  
It is worth noting that the two losses contribute to each other: the advantage weight ensures that the policy constraint is adaptive and selects good samples. Meanwhile, the $Q$ policy gradient guides the policy toward the best action. Based on their
interoperation, we  combine Equation~\ref{eq:adv_bc} with Equation~\ref{eq:td3bc_loss} to optimize the policy concurrently.
This gives the final loss for policy updates
\begin{equation}
    \label{eq:pi_loss}
    L(\psi)=\mathbb{E}_{(s,a)\sim D}[ L_{awbc}(\psi)- \lambda Q_\theta(s,\pi_\psi(s),g)],
\end{equation}
where the scalar $\lambda$ is defined as:
\[
    \lambda = \frac{1}{\frac{1}{N}\sum_{s_i,g,a_i}|Q_\theta(s_i, g, a_i)|}
\]
to balance between RL (in maximizing Q) and imitation (in minimizing the weighted BC term) \cite{TD3BC}.
%The advantage weight ensures the policy model effectively learns optimal actions from very noisy dataset.

\begin{algorithm}
\caption{Q-Advantage policy gradient}\label{alg:DQAPG}
\begin{algorithmic}
 
%\While{training}
    \REPEAT
        \STATE Sample the transitions $\zeta$ and $\zeta_{aug}$ using Alg.~1.
        \STATE Train V-function $V_{\omega}$ by minimizing Eq.~\ref{eq:v_adv}.
        \STATE Train Q-function $Q_{\theta}$ by minimizing Eq.~\ref{eq:q_adv}.
        \STATE Optimize policy $\pi_{\psi}$ by minimizing Eq.~\ref{eq:pi_loss}.
    \UNTIL{Training is stopped}
\end{algorithmic}
\end{algorithm}

\begin{figure*}[h]
    \centering
    \includegraphics[width=.98\textwidth]{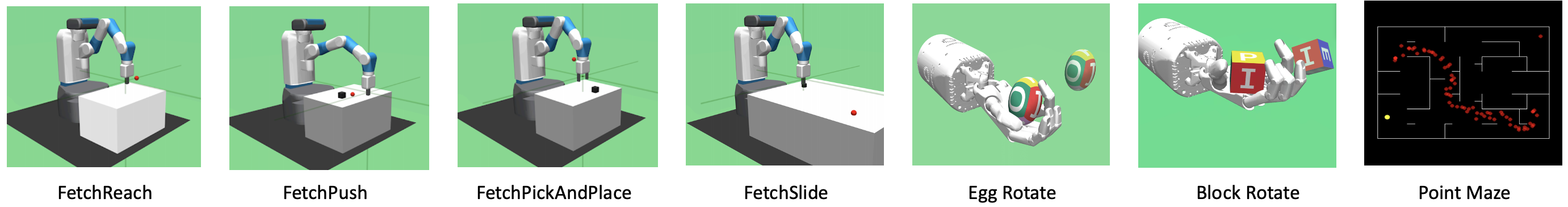}
    \caption{The goal-conditioned tasks selected for experiments in this work. }
    \label{fig:toy-maze}
\end{figure*}

\begin{table*}[t]
\caption{Comparison of offline RL objectives. Note that function $w$ denotes the weighting function in WGCSL, $f'_\star$ denotes the f-divergence function in GoFAR,  $f$ denoted a monotonically increasing function such as exp, and $\sigma$ is the perturbation network in BCQ.
%AdRIL is the latest IL method and reporting SotA performance. 
} 
\label{tab:algo_comp}
\begin{center}
\renewcommand{\arraystretch}{1.4}
\resizebox{1\linewidth}{!}{
  %\begin{tabular}{lSSSSSS ss}
  \begin{tabular}{lcccc}
    \toprule
    \textbf{Offline goal-conditioned RL methods } &
      {Policy objective}&
      {Policy gradient}&
      {Weighted Regression}
     
      \\
     % & Imitation return & True return  & Imitation return & True return & Imitation return & True return & Imitation return & True return & Imitation return & True return\\
    \midrule

GCSL~\cite{GCSL} & $\min_{\pi_\psi} \mathbb{E}_{(s,a,g)\sim D_{relabel}}[-\log \pi_\psi(a|s, g)]$ & \xmark & \xmark\\

WGCSL~\cite{WGCSL} & $\min_{\pi_\psi} \mathbb{E}_{(s_t,a_t,g)\sim D_{relabel}}[-w(A(s_t,g,a_t), t) \cdot \log \pi_\psi(a_t|s_t, g)]$ & \xmark & \cmark\\

GoFAR~\cite{GoFAR} & $\min_{\pi_\psi} \mathbb{E}_{(s,a,g)\sim D} [f'_\star(R(s;g)+\gamma\mathcal{T}\mathcal{V}^*(s,a;g) - \mathcal{V}(s;g))\log \pi_\psi(a|s, g)]$ & \xmark & \cmark\\

AM~\cite{AM} & $\pi(a|s,g)= \underset{a\sim D}{\argmax} Q^*(s,g,a)$ & \xmark & \xmark\\

 \midrule
{\textbf{Offline RL methods }} &
      {Policy objective}&
      {Policy gradient}&
      {Weighted Regression}&
     
      \\
 \midrule

AWR,AWAC,~\cite{AWR,AWAC} & $\min_{\pi_\psi} \mathbb{E}_{(s,a,g)\sim D}[-\exp (A^*(s,g,a)) \cdot \log \pi_\psi(a|s, g)]$ & \xmark & \cmark
\\

%AWR,AWAC,~\cite{AWR,AWAC} & $\min_{\pi_\psi} \mathbb{E}_{(s,a,g)\sim D}[-f(A^*(s,g,a)) \cdot \log \pi_\psi(a|s, g)]$ & \xmark & \cmark x\\
CRR~\cite{CRR}
& $\min_{\pi_\psi} \mathbb{E}_{(s,a,g)\sim D}[-f(Q^*(s,g,a)) \cdot \log \pi_\psi(a|s, g)]$ & \xmark & \cmark
\\

CQL~\cite{CQL}  & $\min_{\pi_\psi}\mathbb{E}_{{s,g}\sim D, a~\sim\pi_\psi} [- \log \pi_\psi(a|s,g) Q(s,g,a)] $ & \cmark & \xmark\\

Onestep-RL, IQL~\cite{onestep, IQL}  & $\min_{\pi_\psi} \mathbb{E}_{(s,a,g)\sim D}[- \log \pi_\psi(a|s, g) Q^{\pi_\psi}(s,g,a)]$ 

%Onestep-RL, IQL & $\min_{\pi_\psi} \mathbb{E}_{(s,a,g)\sim D}[-f(A^*(s,g,a)) \cdot \log \pi_\psi(a|s, g)]$ & \xmark & \cmark x\\

%\\ \textbf{or} $\min_{\pi_\psi}\mathbb{E}_{{s,g}\sim D, a~\sim\pi_\psi} [- \log \pi_\psi(a|s,g) Q^{\pi_\psi}(s,g,a)] $} 
& \cmark & \xmark\\

BCQ ~\cite{BCQ}& $\pi(a|s,g)= \underset{{a\sim[\vae(s,g)+\sigma_\phi(s,g,a)]}}{\argmax}  Q^*(s,g,a)$ & \xmark & \xmark\\

TD3BC ~\cite{TD3BC} & $\min_{\pi_\psi}\mathbb{E}_{(s,a,g)\sim D}[\lVert\pi_\psi(s,g)- a{\rVert}_2 -Q(s,\pi_\psi(s,g),g)$ & \cmark & \xmark\\
\midrule

\textbf{Ours} & $\min_{\pi_\psi}\mathbb{E}_{(s,a,g)\sim D}[f(A(s,g,a))\lVert\pi_\psi(s,g) - a {\rVert}_2 -Q(s,\pi_\psi(s,g),g)$ & \cmark & \cmark\\
    \bottomrule
  \end{tabular}}
\end{center}
\label{tab:all_comparisons}

\end{table*}

\textbf{Algorithm summary.\quad}
The pseudo-code of the optimization steps of DQAPG is in Algorithm~\ref{alg:DQAPG}. The two main steps are i) sample and augment goal-conditioned tuples using Algorithm~\ref{alg:goal-swap}, and
ii) then use the mixed augmented tuples $(\zeta, \zeta_{aug})$ in off-policy Q-Advantage policy gradient optimization. The weighted behavioral cloning $L(\psi)_{awbc}$ filters out the non-optimal goal-conditioned actions, 
%the dynamic programming nature of TD-learning (Eq.\ref{eq:q_adv}) connects all possible reachable goals 
and $Q(s,g,a)$ provides deterministic policy gradient to optimize the policy $\pi_\psi$.

\subsection{Connection to prior works}
\label{subsec:connections}
Table~\ref{tab:algo_comp} compares the proposed policy optimization objective to the prior works. For comparison, four state-of-the-art offline GCRL
methods, and seven general offline conditioned RL methods were selected.

Table~\ref{tab:algo_comp} shows that most of the prior GCRL methods are weighted regression-based approaches, which form an objective similar to Eq.~\ref{eq:adv_bc}. However, out of these GCSL~\cite{GCSL} and WGCSL~\cite{WGCSL} can use only trajectories of successful examples, GoFAR~\cite{GoFAR} require that the offline dataset covers all reachable goals, and finally AM~\cite{AM}
fails on noisy trajectories. In contrast, the proposed method proposed avoids all these shortcomings.
%As is discussed above, purely advantage weighted regression can be non-optimal, especially for a large noisy dataset. Com 
%As previously introduced, the key for offline RL is to control distribution shift by learning enforce the policy generate actions within the dataset.  
 %It is worth noting that purely apply conservative 

On the other hand, our work shares the idea of the prior offline RL methods AWAC~\cite{AWAC}, CRR~\cite{CRR}, and AWR~\cite{AWR}. To properly acknowledge these prior works, we conclude that the proposed method takes advantage of both types of methods by jointly using the deterministic policy gradient optimization and the weighted regression.
For the same reason, many of the above methods, in particular, AWAC, CRR, WGCSL, and GoFAR, can be integrated into our model.
% PERHAPS TOO MUCH DETAILS
%by using the deterministic policy gradient or adding the weighted regression loss (e.g, TD3BC).
 
 %We can modify their policy optimization objective similar to $L(\psi)_{awbc}$  by replacing their stochastic policies with the deterministic policy. Compared with TD3BC, our method can be considered as an improved TD3BC. TD3BC uses a fixed behavior cloning regularization term, which potentially makes the policy behaves too conservatively, while our method has an adaptive policy constraint that makes the learned policy less pessimistic.

% As is discussed in Sec.\ref{subsec:offline_rl_related}, 

\begin{figure*}[h]
    \centering
    \includegraphics[width=.8\textwidth]{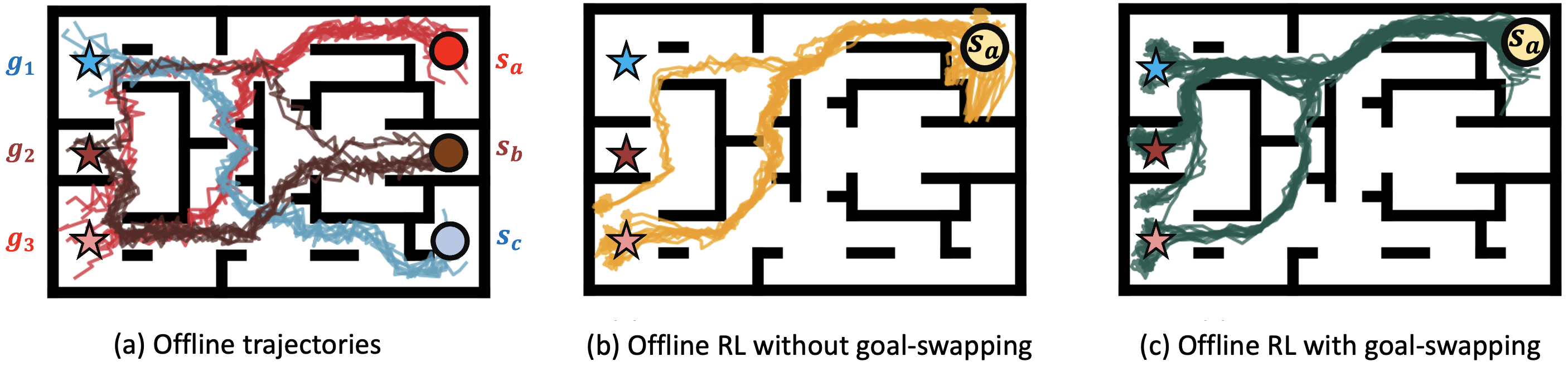}
    \caption{A PointMaze (toy) example.
    DQAPG was used and during the test time the agent always starts at $s_0$, and goal is randomply picked from $[g_1, g_2, g_3]$. (a) 3 types of offline trajectories: $s_a$ to $g_3$, $s_b$ to $g_2$, and $s_c$ to $g_1$. (b) offline RL performance without data augmentation (only one goal is achieved due to serious overfitting). (c) offline RL performance with goal-swapping augmentation.}%$\tau^a$ and $\tau^b$ have mutual visited state $s_t$. The

    \label{fig:goal-swap-comparison}
\end{figure*}

% TABLE about Method comparison
\begin{table*}[t]
\caption{Evaluation of the goal-swapping augmentation in offline RL. The tested methods were trained without (*method*) and with (*method*-aug) the proposed augmentation. The performance metric is the average cumulative reward over 50 random episodes.
%and We compare the original algorithms and their data-augmentation variants to verify the effectiveness of goal-swapping.  "\{Algorithm name\}"-aug indicates the algorithm's data-augmented variant.
\colorbox{pink}{\enspace\enspace\enspace} indicates that the  augmented variant outperforms the original algorithm according to the t-test with p-value $< 0.05$. 
} 
\label{tab:aug_compare}
\begin{center}
\resizebox{.95\linewidth}{!}{
  %\begin{tabular}{lSSSSSS ss}
  \begin{tabular}{lcccccccccccc}
    \toprule
     {\textbf{Datasize-small}} &
      {FetchPush}&
      {FetchSlide} &
      {FetchPick} &
      {FetchReach}&
      {HandBlockZ}&
      {HandBlockParallel}&
      {HandBlockXYZ} &
      {HandEggRotate}&
      {PointMaze} &\\
     % & Imitation return & True return  & Imitation return & True return & Imitation return & True return & Imitation return & True return & Imitation return & True return\\
    \midrule

DQAPG-aug & \colorbox{pink}{$\mathbf{29.06 \pm 14.94}$}  & $2.0 \pm 4.57$ & \colorbox{pink}{$\mathbf{32.39 \pm 12.45}$} & $48.1 \pm 0.94$ & $31.69 \pm 35.88$ & $11.68 \pm 19.8$ & $20.49 \pm 28.35$ & \colorbox{pink}{$\mathbf{36.13 \pm 33.9}$} & \colorbox{pink}{$\mathbf{58.15\pm19.83}$}  \\ % True   

DQAPG & $25.33 \pm 17.71$ & $1.25 \pm 3.05$ & $30.04 \pm 14.79$ & $48.14 \pm 0.88$ & $35.04 \pm 35.85$ & $16.8 \pm 25.47$ & $22.66 \pm 28.27$ & $34.08 \pm 37.28$ & $34.74\pm30.16$  \\ % False  

%\midrule
%awac-adv & $29.06 \pm 14.94$ & $2.0 \pm 4.57$ & $32.39 \pm 12.45$ & $48.24 \pm 0.86$ & $18.78 \pm 33.25$ & $0.0 \pm 0.0$ & $7.43 \pm 20.64$ & $18.64 \pm 28.04$ \  \\ % True  

%awac-adv & $22.35 \pm 17.92$ & $1.77 \pm 3.67$ & $30.95 \pm 13.0$ & $48.26 \pm 0.86$ & $12.36 \pm 29.81$ & $0.0 \pm 0.0$ & $12.26 \pm 26.22$ & $17.6 \pm 32.06$ \  \\ % False  

\midrule
TD3BC-aug & \colorbox{pink}{$\mathbf{17.5 \pm 15.47}$} & $0.43 \pm 1.48$ & $29.68 \pm 11.53$ & $47.82 \pm 2.28$ & \colorbox{pink}{$\mathbf{19.1 \pm 31.03}$} & $2.61 \pm 10.14$ & $13.24 \pm 25.24$ & \colorbox{pink}{$\mathbf{33.27 \pm 36.53}$} & \colorbox{pink}{$\mathbf{51.26 \pm 14.60}$}  \\ % True  

TD3BC  & $12.94 \pm 17.07$ & $1.14 \pm 4.78$ & $28.86 \pm 13.35$ & $47.92 \pm 1.07$ & $17.25 \pm 30.98$ & $3.77 \pm 13.4$ & $12.76 \pm 24.44$ & $29.58 \pm 36.62$ & $28.00\pm26.82$  \\ % False  

\midrule
AWAC-aug &  \colorbox{pink}{$\mathbf{17.9 \pm 15.82}$} & $1.54 \pm 3.79$ & $12.49 \pm 13.79$ & $42.94 \pm 4.02$ & \colorbox{pink}{$\mathbf{13.86 \pm 18.74}$} & $0.26 \pm 1.41$ & \colorbox{pink}{$\mathbf{1.59 \pm 7.64}$} & \colorbox{pink}{$\mathbf{7.58 \pm 10.03}$} & \colorbox{pink}{$\mathbf{43.60\pm25.44}$}  \\ % True  

AWAC  & $15.11 \pm 16.61$ & $0.94 \pm 2.3$ & $12.19 \pm 13.39$ & $43.53 \pm 3.56$ & $0.93 \pm 4.57$ & $0.0 \pm 0.0$ & $0.5 \pm 3.01$ & $0.02 \pm 0.18$ & $32.12\pm29.78$  \\ % False  

\midrule

IQL-aug & $3.5 \pm 12.24$ & $0.54 \pm 4.56$ & $1.26 \pm 7.67$ & $16.12 \pm 17.01$ & $1.45 \pm 9.01$ & $0.0 \pm 0.0$ & $0.34 \pm 2.98$ & $0.0 \pm 0.0$  & \colorbox{pink}{$\mathbf{20.72\pm18.85}$} \\ % True  

IQL  & $3.73 \pm 12.92$ & $0.59 \pm 4.55$ & $1.32 \pm 6.81$ & $20.2 \pm 18.2$ & $1.43 \pm 9.01$ & $0.0 \pm 0.0$ & $0.18 \pm 1.29$ & $0.0 \pm 0.0$  & $14.45\pm20.83$ \\ % False  

\midrule
Onestep-RL-aug & \colorbox{pink}{$\mathbf{27.38 \pm 15.3}$} & $0.22 \pm 0.99$ & \colorbox{pink}{$\mathbf{21.18 \pm 17.51}$} & \colorbox{pink}{$\mathbf{48.03 \pm 1.44}$} & \colorbox{pink}{$\mathbf{1.81 \pm 6.07}$} & $0.42 \pm 4.63$ & $0.09 \pm 0.49$ & $0.59 \pm 1.86$ & $0.73 \pm 1.03$  \\ % True  

Onestep-RL  & $5.52 \pm 11.94$ & $1.12 \pm 2.34$ & $9.64 \pm 14.86$ & $45.9 \pm 9.65$ & $0.23 \pm 1.61$ & $0.0 \pm 0.0$ & $0.06 \pm 0.63$ & $0.01 \pm 0.09$ & $0.23 \pm 1.45$  \\ % False  

\midrule
CRR-aug & \colorbox{pink}{$\mathbf{21.67 \pm 15.52}$} & $1.47 \pm 3.53$ & $16.73 \pm 14.18$ & $45.43 \pm 2.18$ & $0.34 \pm 1.38$ & $0.0 \pm 0.0$ & $0.25 \pm 1.7$ & $0.18 \pm 0.95$ & \colorbox{pink}{$\mathbf{44.83\pm32.28}$}  \\ % True  

CRR  & $16.5 \pm 14.7$ & $2.22 \pm 5.67$ & $15.98 \pm 14.26$ & $45.65 \pm 2.2$ & $0.73 \pm 5.0$ & $0.0 \pm 0.0$ & $0.4 \pm 2.18$ & $0.1 \pm 1.07$  & $26.32\pm31.77$   \\ % False  

\midrule
CQL-aug & \colorbox{pink}{$\mathbf{31.18 \pm 11.35}$} & \colorbox{pink}{$\mathbf{4.97 \pm 6.97}$} & \colorbox{pink}{$\mathbf{24.65 \pm 12.54}$} & $44.6 \pm 2.18$ & $0.17 \pm 0.79$ & $0.0 \pm 0.0$ & $0.02 \pm 0.18$ & $0.0 \pm 0.0$ & \colorbox{pink}{$\mathbf{16.56 \pm 28.14}$} \\ % True  

CQL  & $22.02 \pm 17.24$ & $1.92 \pm 3.25$ & $19.18 \pm 13.74$ & $44.15 \pm 2.36$ & $0.14 \pm 0.66$ & $0.0 \pm 0.0$ & $0.04 \pm 0.32$ & $0.0 \pm 0.0$ & $7.76\pm18.42$  \\ % False  

\bottomrule
  \end{tabular}}
\end{center}
\label{tab:all_comparisons}
%\vspace{-15pt}
\end{table*}
% TABLE ENDS

%Comapred with the offline goal-conditioned RL methods,  

\section{Experiments}
\label{subsec:experiment}
The experiments in this section were designed to verify our main claims:
1) the goal-swapping augmentation helps to learn a more general policy;
2) the DQAPG improves the performance of goal-conditioned RL (GCRL); and
3) the method learns a working general policy from offline data.
%In this section, experiments to answer the following questions: 1) Can the goal-swapping augmentation contribute to the generalizability? 2) Can our DQAPG method contribute to the offline GCRL performance? 3)  Can the proposed pipeline learn a generalized goal-conditioned policy from the offline dataset?  

\textbf{Tasks.\quad} 
Nine goal-conditioned tasks from~\cite{Tasks} were selected for the experiments.
The tasks include the simple PointMaze task, four fetching manipulation tasks (FetchReach, FetchPickAndPlace, FetchPush, FetchSlide), and four dexterous in-hand manipulation tasks (HandBlock-Z, HandBlock-XYZ, HandBlock-Parallel, HandEgg). In the fetching tasks, the virtual robot should move an object to a specific position in the virtual space. In the four dexterous in-hand manipulation tasks, the agent is asked to rotate the object to a specific pose. The offline dataset for each task is a mixture of 500 expert trajectories and 2,000 random trajectories. The fetching tasks and datasets are taken from ~\cite{WGCSL}. The expert trajectories for the in-hand manipulation tasks are generated similarly to~\cite{MRN}.
For more details of the tasks, we refer to Appendix~\ref{subsec:dataset}.
%In our simulation experiments, we consider six distinct environments. They in- clude four robot manipulation environments[39]: FetchReach, FetchPickAndPlace, FetchPush, and FetchSlide, and 4 dexterous manipulation environments: HandBlock-Z, HandBlock-XYZ, HandBlock-Parallel and HandEgg[39]. All tasks use sparse reward, fer source (middle) and target (right) domains. and their respective goal distributions are defined over valid configurations in either the robot space or object space, depending on whether the task involves object manipulation. The offline dataset for each task is collected by either a random policy or a mixture of 2000 random policy trajectories and 1000 expert policy trajectories, depending on whether random data provides enough coverage of the desired goal distribution. The first five tasks and their datasets are taken from [????]. 

\textbf{Baselines.\quad} 
The selected baseline methods are listed in Table.\ref{tab:algo_comp}. The HER ratio of 0.5 was used with all methods in all experiments.
We trained each method for 10 seeds, and each training run uses 500k updates. The mini-batch size is 512. Complete architecture, hyper-parameter table, and additional training details are provided in Appendix~\ref{subsec:exp_details}.

We used the cumulative test rewards from the environments as the performance metric in all experiments.

% TABLE about Method comparison
\begin{table*}[t]
\caption{DQAPG vs. the strong baselines for offline RL. No goal-swapping augmentation is used in this experiment. \colorbox{mycolor}{\enspace\enspace} indicates the best results according to the t-test with p-value $< 0.05$. 
} 
\label{tab:DQAPG_eval}
\begin{center}
\resizebox{.95\linewidth}{!}{
  %\begin{tabular}{lSSSSSS ss}
  \begin{tabular}{lcccccccccccccccc}
    \toprule
     {\textbf{}} &
      \textbf{DQAPG}&
      {TD3BC} &
      {AWAC} &
      {IQL} &
      {Onestep-RL} &
      {CRR} &
      {BCQ} &
      {CQL} &\\
     % & Imitation return & True return  & Imitation return & True return & Imitation return & True return & Imitation return & True return & Imitation return & True return\\
    \midrule

%\textbf{small dataset}
%\midrule

FetchPush & \colorbox{mycolor}{$25.33 \pm 17.71$}  & $15.46\pm 17.59$ & $15.11 \pm 16.61$ & $3.8 \pm 12.72$ & $4.26 \pm 11.33$ & $16.5 \pm 14.7$ & $3.78 \pm 13.02$ & $22.78 \pm 15.77$ \  \\ % False  
FetchSlide  &  \colorbox{mycolor}{$2.21 \pm 5.6$} & $0.84 \pm 2.43$   & $0.94 \pm 2.3$  & $0.53 \pm 4.51$ & $0.62 \pm 1.66$ &  \colorbox{mycolor}{$2.22 \pm 5.67$} & $0.44 \pm 4.47$ &  \colorbox{mycolor}{$1.95 \pm 2.93$} \  \\ % False  
FetchPick &  \colorbox{mycolor}{$30.04 \pm 14.79$}  &  \colorbox{mycolor}{$29.1 \pm 12.67$}  & $12.19 \pm 13.39$ & $1.67 \pm 7.52$ & $6.58 \pm 12.3$ & $15.98 \pm 14.26$ & $1.2 \pm 7.65$ & $21.02 \pm 13.24$ \  \\ % False  
FetchReach  &  \colorbox{mycolor}{$48.25 \pm 0.86$} &  \colorbox{mycolor}{$47.95 \pm 1.06$}   & $43.53 \pm 3.56$ & $21.76 \pm 17.48$ & $45.6 \pm 10.14$ & $45.65 \pm 2.2$ & $6.62 \pm 12.62$ & $44.54 \pm 2.44$ \  \\ % False  
HandBlock-Z &  \colorbox{mycolor}{$35.04 \pm 35.85$}  & $16.92 \pm 30.81$   & $0.93 \pm 4.57$  & $1.56 \pm 8.53$ & $0.22 \pm 1.11$ & $0.73 \pm 5.0$ & $1.88 \pm 12.85$ & $0.14 \pm 0.66$ \  \\ % False  
HandBlock-Parallel &  \colorbox{mycolor}{$16.8 \pm 25.47$}  & $3.5 \pm 13.72$   & $0.0 \pm 0.0$ & $0.22 \pm 2.32$ & $0.0 \pm 0.0$ & $0.0 \pm 0.0$ & $0.0 \pm 0.0$ & $0.0 \pm 0.0$ \  \\ % False  
HandBlock-XYZ &  \colorbox{mycolor}{$22.66 \pm 28.27$}  & $10.68 \pm 22.71$  & $0.5 \pm 3.01$  & $0.02 \pm 0.2$ & $0.02 \pm 0.18$ & $0.4 \pm 2.18$ & $0.78 \pm 8.73$ & $0.0 \pm 0.0$ \  \\ % False 
HandEggRotate &  \colorbox{mycolor}{$35.08 \pm 37.28$}  & $29.58 \pm 36.62$   & $0.02 \pm 0.18$ & $0.0 \pm 0.0$ & $0.01 \pm 0.09$ & $0.1 \pm 1.07$ & $0.0 \pm 0.0$ & $0.0 \pm 0.0$ \  \\ % False  

PointMaze   &  \colorbox{mycolor}{$34.74\pm30.16$} & $28.00\pm26.82$ & $32.12\pm29.78$ & $14.45\pm20.83$ & $0.0\pm0.0$ & $26.32\pm31.77$ & $0.0\pm0.0$ & $7.76\pm18.42$\\
%Halfcheetah  & $1960.2 \pm 49~~$ & 1950.2 $\pm$ 50~ & 1965.3 $\pm$ 36~ & \underline{1974.4 $\pm$ 47~} & $\mathbf{2083.6 \pm 19~}$ &  $ -550.00 \pm 179$ & $-1237 \pm 234$ & \underline{1902.3 $\pm$ 108}\\
    \bottomrule
  \end{tabular}}
\end{center}
\label{tab:all_comparisons}
\end{table*}
% TABLE ENDS

% TABLE about Method comparison
\begin{table*}[t]
\caption{
% Overall offline GCRL performance comparisons. The \textbf{bold font} names indicate that goal-swapping  is used. The \textit{italic font} names indicate the methods cannot use goal-swapping. \colorbox{mycolor}{    } highlights the best results according to the t-test with p-value $< 0.05$. 
Overall comparisons. The methods using goal-swapping augmentation are indicated in \textbf{bold}, and the methods in \textit{italic} cannot use goal-swapping augmentation.  \colorbox{mycolor}{    } highlights the best results according to the t-test with p-value $< 0.05$. 
} 
\begin{center}
\resizebox{1.0\linewidth}{!}{
  %\begin{tabular}{lSSSSSS ss}
  \begin{tabular}{lcccccccccccccccc}
    \toprule
     {\textbf{}} &
      {\textbf{DQAPG}}&
      \textbf{TD3+BC} &
      \textbf{AM} &
      \textit{WGCSL}&
      \textit{GCSL}&
      \textit{GoFAR}&
      \textbf{AWAC} &
      \textbf{IQL} &
      \textbf{Onestep-RL} &
      \textbf{CRR} &
      \textbf{BCQ} &
      \textbf{CQL} &\\
     % & Imitation return & True return  & Imitation return & True return & Imitation return & True return & Imitation return & True return & Imitation return & True return\\
    \midrule

FetchPush  & \colorbox{mycolor}{$30.36 \pm 15.23$} & $12.23 \pm 15.63$ & $19.29 \pm 16.33$ & $10.14 \pm 16.19$ & $8.98 \pm 15.37$ & $12.42 \pm 16.58$ & $8.05 \pm 16.01$ & $3.66 \pm 12.92$ & $11.96 \pm 14.67$ & $21.67 \pm 15.52$ & $3.82 \pm 13.1$ & \colorbox{mycolor}{$31.05 \pm 11.41$} \  \\ % True  
FetchSlide  & $2.09 \pm 4.8$ & $0.95 \pm 4.94$ & \colorbox{mycolor}{$7.45 \pm 11.41$} & $2.03 \pm 7.03$ & $0.61 \pm 1.79$ & $1.75 \pm 5.73$  & $1.7 \pm 6.03$ & $0.29 \pm 1.52$ & $2.51 \pm 5.02$ & $1.47 \pm 3.53$ & $0.44 \pm 4.47$ & $4.94 \pm 7.6$ \  \\ % True  
FetchPick & \colorbox{mycolor}{$33.38 \pm 10.86$} & $27.34 \pm 12.65$ & $18.55 \pm 19.08$ & $6.56 \pm 11.46$ & $7.38 \pm 12.07$ & $16.39 \pm 16.17$
 & $11.48 \pm 13.98$ & $0.88 \pm 5.82$ & $18.35 \pm 18.74$ & $16.73 \pm 14.18$ & $1.2 \pm 7.65$ & $25.4 \pm 12.38$ \  \\ % True  

FetchReach & \colorbox{mycolor}{$48.24 \pm 0.85$} & \colorbox{mycolor}{$47.81 \pm 1.14$} & \colorbox{mycolor}{$48.24 \pm 0.86$} & $36.96 \pm 9.11$ & $36.1 \pm 9.35$ & $45.15 \pm 5.51$
 & $40.35 \pm 11.39$ & $17.21 \pm 17.54$ & $47.2 \pm 6.43$ & $45.43 \pm 2.18$ & $6.41 \pm 12.49$ & $44.38 \pm 2.51$ \  \\ % True  
HandBlock-Z & \colorbox{mycolor}{$31.69 \pm 35.88$}  & $18.02 \pm 30.14$ & $0.65 \pm 6.78$ & $2.78 \pm 11.34$ & $3.08 \pm 11.37$ & $5.67 \pm 16.17$
  & $14.38 \pm 24.07$ & $1.34 \pm 8.9$ & $1.31 \pm 5.23$ & $0.34 \pm 1.38$ & $2.26 \pm 12.75$ & $0.17 \pm 0.79$ \  \\ % True  
HandBlock-Parallel & \colorbox{mycolor}{$11.68 \pm 19.8$}  & $2.04 \pm 8.04$ & $0.0 \pm 0.0$ & $0.02 \pm 0.18$ & $0.0 \pm 0.0$ & $0.01 \pm 0.09$  & $0.27 \pm 2.03$ & $0.0 \pm 0.0$ & $0.01 \pm 0.09$ & $0.0 \pm 0.0$ & $0.0 \pm 0.0$ & $0.0 \pm 0.0$ \  \\ % True  
HandBlock-XYZ & \colorbox{mycolor}{$20.49 \pm 28.35$}  & $12.38 \pm 22.54$ & $0.04 \pm 0.37$ & $1.53 \pm 7.89$ & $0.34 \pm 1.82$ & $0.95 \pm 5.01$
  & $0.03 \pm 0.25$ & $0.71 \pm 5.47$ & $0.12 \pm 0.8$ & $0.25 \pm 1.7$ & $0.73 \pm 6.96$ & $0.05 \pm 0.38$ \  \\ % True  
HandEggRotate & \colorbox{mycolor}{$36.13 \pm 33.9$}  & $33.27 \pm 36.53$ & $0.0 \pm 0.0$ & $0.08 \pm 0.89$ & $1.18 \pm 6.41$ & $4.7 \pm 14.81$ & $8.42 \pm 22.24$ & $0.0 \pm 0.0$ & $0.59 \pm 1.86$ & $0.18 \pm 0.95$ & $0.0 \pm 0.0$ & $0.0 \pm 0.0$  \  \\ % True  

PointMaze   & \colorbox{mycolor}{$58.15\pm19.83$} & $51.26\pm14.60$ &$0.24\pm1.59$ &$29.31\pm21.78$ & $26.78\pm22.95$ & $28.26\pm32.25$ & $43.60\pm25.44$ & $20.72\pm18.85$ & $0.0\pm0.0$ & $44.83\pm32.28$ & $0.0\pm0.0$            & $16.56\pm28.14$\\
%\textbf{small dataset}
%\midrule

%Halfcheetah  & $1960.2 \pm 49~~$ & 1950.2 $\pm$ 50~ & 1965.3 $\pm$ 36~ & \underline{1974.4 $\pm$ 47~} & $\mathbf{2083.6 \pm 19~}$ &  $ -550.00 \pm 179$ & $-1237 \pm 234$ & \underline{1902.3 $\pm$ 108}\\
    \bottomrule
 \end{tabular}}
\end{center}
\label{tab:all_comparisons}
\end{table*}
% TABLE ENDS

%\subsection{Performance evaluation}

\textbf{Goal-swapping experiment.\quad}
%\paragraph{Effectiveness of goal-augmentation}
At first, for demonstration purposes, a qualitative demonstration of the goal-swapping augmentation is presented in Figure~\ref{fig:goal-swap-comparison}. 

 In this task, the test cases are different from the offline training data, and therefore it exemplifies the generalization power of the proposed augmentation technique. The offline training data was generated, by storing 10 trajectories from the three fixed paths: $s_a\rightarrow g_3$,  $s_b\rightarrow g_2$, and  $s_c\rightarrow g_1$ (visualized by different colors in Figure~\ref{fig:goal-swap-comparison}(a)). During testing, however, the starting point is fixed and goals were randomly sampled, thus representing unseen starting points and goal combinations. Two versions of the proposed DQAPG method were trained, one
 with the goal-swapping augmentation ($\star$-aug) and another without the augmentation. Each variant was tested using
 50 random episodes.
%Figure~\ref{fig:goal-swap-comparison} (b) visualize the trajectories of DQAPG without goal-swapping, Figure~\ref{fig:goal-swap-comparison} (b) visualizes the performance of DQAPG with goal-swapping augmentation.

The qualitative test results are shown in Figure~\ref{fig:goal-swap-comparison}(b) without augmentation and in Figure~\ref{fig:goal-swap-comparison}(c) with augmentation. These results indeed verify that the proposed augmentation technique is able to provide a more general policy that can generalize beyond its training data.
%Appendix~\ref{subsec:pointmaze_fig} provides additional results.

%Figure~\ref{fig:toy-maze}

 %In this experiment, we demonstrate how goal-swapping can significantly improve the offline goal-conditioned RL learning with limited dataset. we demonstrate results of  goal-swapping augmentation in a simple PointMaze environment.

% We also conduct the experiment in \textbf{FetchReach} task, with only \textbf{60 random trajectories} given as offline RL dataset. 

%\paragraph{Effectiveness of goal-swapping augmentation} 
%As the goal-swapping augmentation has been qualitatively evaluated above, we first present the quantitative evaluation to study its effectiveness and limitations in this section.
Next, we needed to verify that the augmentation technique itself generalizes to other tasks as well. The results for multiple methods and all nine task are shown in Table~\ref{tab:aug_compare}.
The goal-swapping can benefit only methods using TD-learning, and therefore the following were selected for comparison: TD3+BC, AWAC, CRR, IQL, Onestep-RL, CQL, and our DQAPG.

This experiment provides an important finding, the goal-swapping augmentation does not provide significant negative effect if the task does not benefit from it, but in many cases it provides a significant performance boost. The tasks that benefit significantly from the augmentation were FetchPush and PointMaze. The tasks that obtained smaller improvements were FetchPick, HandBlockZ, and HandEggRotate. The rest of the tasks were more or less intact.

On the method side, AWAC seems to benefit the most from the augmentation (five tasks in total). All offline RL methods, except IQL, as well benefit from the augmentation. It seems that none of the methods benefit from the augmentation in two tasks, FetchSlide and HandBlockParallel. However, since the augmentation technique does not seem to have negative impact it is safe to be used.

% THESE ARE PERHAPS LESS INTERESTING DETAILS /JONIK
%which the nature of the tasks can explain. The FetchSlide task is generally more challenging as the robot only hits the object once, which generates sparse trajectories. For the HandBlockParallel task, the success indicator function has a more strict threshold than other in-hand manipulation tasks, which may affect the evaluation results. All methods can perfectly accomplish the FeachReach task regardless of whether the data is augmented. Overall, the qualitative evaluation shows how goal-swapping effectively connects goals across trajectories, and the quantitative evaluation verifies its effectiveness on some goal-conditioned tasks. 

\textbf{DQAPG performance experiment.\quad}
%We design this experiment to evaluate our proposed deterministic Q-Advantage policy gradient loss.
It is essential to verify that the proposed DQAPG optimization objective is general and not problem specific.
% THIS I DONT UNDERSTAND /JONIK
%As discussed in Section~\ref{subsec:connections}, DQAPG should be compared with TD3BC and the weighted regression-based methods AWAC and CRR, and multiple others.
The offline RL results are summarized in Table~\ref{tab:DQAPG_eval}. Overall, the DQAPG is the best in all nine tasks, and shares the best place in four tasks. 
%
%ance is strong in most of the tasks. Based on the evaluation results,  CQL and CRR have similar performance as DQAPG on the FetchSlide task, and TD3BC performs as well as DQAPG on FetchPick and FetchReach tasks.
It is worth noting that the basic TD3BC remains as a strong general baseline for offline goal-conditioned RL
despite of its simplicity as compared to others.
% despite being slightly outperformed by DQAPG. 
% In general, our proposed DQAPG has the best performance across all goal-conditioned tasks. It is verified that techniques used in DQAPG are effective in generating better policies than current prior methods in offline goal-conditioned RL settings.
These results verify that the DQAPG's objective function
suits well for offline goal-conditioned RL. This setting is important in robotics where offline data provides superior sample efficiency and safety as compared to online learning with a real robot.
%In general, DQAPG achieves the best performance across all goal-conditioned tasks, indicating its effectiveness in generating better policies than prior methods in the offline goal-conditioned RL setting.

\textbf{Offline goal-conditioned RL comparison.\quad} 
In Table~\ref{tab:all_comparisons}, are the results for the same tasks but using also the goal-swapping augmentation. The proposed augmentation technique indeed improves the performance of all methods (results without augmentation included for comparison). The goal-swapping augmentation was implemented on all TD-learning-based methods except GCSL, WGCSL, and GoFAR due to their own competing optimization objectives. 
Overall, the proposed method again performs the best.
However, for the FetchSlide task, AM outperforms all other methods.
It is worth noting that although TD3BC is a basic baseline for offline RL, it still outperforms many of the state-of-the-art GCRL methods such as WGCSL and GoFAR.

Almost all baselines failed for the challenging dexterous in-hand manipulation tasks (HandBlock-Parallel, HandBlock-Z, HandBlock-XYZ), for which the proposed method and data augmentation obtained the best performance.

%\paragraph{Affects of expectile loss}
%Expectile regression plays an important role in this method as it tries to estimate the nearly-optimal value functions. We designed the eperiment to study how different expectile parameters $\tau$ affect the offline goal-conditioned RL training. 

\section{Conclusion}
In this work, we proposed a simple yet efficient offline goal-conditioned RL method, DQAPG, and a novel goal-swapping data augmentation technique. DQAPG deals better with noisy offline data than the existing methods, and the data augmentation technique allows offline RL methods to learn general policies beyond the provided offline samples. All findings were verified by extensive experiments on diverse offline learning tasks.
% THIS KIND OF SAYS NOTHING THAT THE PREVIOUS DONT ALREADY SAY
%We evaluate how goal-swapping augmentation enables generalized goal-conditioned policy learning and find its limitations on some tasks where the trajectories are too sparse to connect goals (FetchSlide).  All results were verified by a wide set of goal-conditioned RL tasks, and our method systematically outperforms prior methods. 

%Imitation Q-learning method, CEMD, that learns to solve a control task using expert demonstrations to guide its learning. The expert demonstrations make Imitation Learning more sample efficient than plain Reinforcement Learning (RL), and therefore more suitable for real robot cases. Our main contributions are the tighter bounds for Earth Mover's Distance (EMD) based reward computation. The bounds boost Q-learning to converge faster, to achieve higher performance, and to reduced variance. The results were verified by multiple continuous control tasks where the proposed method systematically outperforms its competitors. All code will be made publicly available.

\section{Limitations and Future work}

The current form of goal-swapping augmentation is random and thus not necessarily efficient. 
More optimal swapping should be investigated. 
Another limitation is that DQAPG can learn only a deterministic policy. Nevertheless, the method can be converted to accept stochastic policies which are optimized using the re-parameterization trick. Another interesting research direction is to find methods to distinguish positive goal-swapping trajectories from negative (unreachable) ones. 
%In the future, it is possible to create a mechanism to remember and prioritize the positive goal-swapping augmentation for experience replay to maximize data utilization.  We are excited to continue to extend this approach in our future works.

% In the unusual situation where you want a paper to appear in the
% references without citing it in the main text, use \nocite
\nocite{langley00}

\bibliography{icml2023}

\begin{thebibliography}{35}
\providecommand{\natexlab}[1]{#1}
\providecommand{\url}[1]{\texttt{#1}}
\expandafter\ifx\csname urlstyle\endcsname\relax
  \providecommand{\doi}[1]{doi: #1}\else
  \providecommand{\doi}{doi: \begingroup \urlstyle{rm}\Url}\fi

\bibitem[Andrychowicz et~al.(2017)Andrychowicz, Wolski, Ray, Schneider, Fong,
  Welinder, McGrew, Tobin, Pieter~Abbeel, and Zaremba]{HER}
Andrychowicz, M., Wolski, F., Ray, A., Schneider, J., Fong, R., Welinder, P.,
  McGrew, B., Tobin, J., Pieter~Abbeel, O., and Zaremba, W.
\newblock Hindsight experience replay.
\newblock \emph{Advances in neural information processing systems}, 30, 2017.

\bibitem[Brandfonbrener et~al.(2021)Brandfonbrener, Whitney, Ranganath, and
  Bruna]{onestep}
Brandfonbrener, D., Whitney, W., Ranganath, R., and Bruna, J.
\newblock Offline rl without off-policy evaluation.
\newblock \emph{Advances in Neural Information Processing Systems},
  34:\penalty0 4933--4946, 2021.

\bibitem[Brunke et~al.(2022)Brunke, Greeff, Hall, Yuan, Zhou, Panerati, and
  Schoellig]{robot1}
Brunke, L., Greeff, M., Hall, A.~W., Yuan, Z., Zhou, S., Panerati, J., and
  Schoellig, A.~P.
\newblock Safe learning in robotics: From learning-based control to safe
  reinforcement learning.
\newblock \emph{Annual Review of Control, Robotics, and Autonomous Systems},
  5:\penalty0 411--444, 2022.

\bibitem[Chane-Sane et~al.(2021)Chane-Sane, Schmid, and Laptev]{gcrl2}
Chane-Sane, E., Schmid, C., and Laptev, I.
\newblock Goal-conditioned reinforcement learning with imagined subgoals.
\newblock In \emph{International Conference on Machine Learning}, pp.\
  1430--1440. PMLR, 2021.

\bibitem[Chebotar et~al.(2021)Chebotar, Hausman, Lu, Xiao, Kalashnikov, Varley,
  Irpan, Eysenbach, Julian, Finn, et~al.]{AM}
Chebotar, Y., Hausman, K., Lu, Y., Xiao, T., Kalashnikov, D., Varley, J.,
  Irpan, A., Eysenbach, B., Julian, R., Finn, C., et~al.
\newblock Actionable models: Unsupervised offline reinforcement learning of
  robotic skills.
\newblock \emph{arXiv preprint arXiv:2104.07749}, 2021.

\bibitem[Frazier \& Riedl(2019)Frazier and Riedl]{game3}
Frazier, S. and Riedl, M.
\newblock Improving deep reinforcement learning in minecraft with action
  advice.
\newblock In \emph{Proceedings of the AAAI conference on artificial
  intelligence and interactive digital entertainment}, volume~15, pp.\
  146--152, 2019.

\bibitem[Fu et~al.(2021)Fu, Kumar, Nachum, Tucker, and Levine]{D4RL}
Fu, J., Kumar, A., Nachum, O., Tucker, G., and Levine, S.
\newblock D4{\{}rl{\}}: Datasets for deep data-driven reinforcement learning,
  2021.
\newblock URL \url{https://openreview.net/forum?id=px0-N3_KjA}.

\bibitem[Fujimoto \& Gu(2021)Fujimoto and Gu]{TD3BC}
Fujimoto, S. and Gu, S.~S.
\newblock A minimalist approach to offline reinforcement learning.
\newblock \emph{Advances in neural information processing systems},
  34:\penalty0 20132--20145, 2021.

\bibitem[Fujimoto et~al.(2018)Fujimoto, Hoof, and Meger]{TD3}
Fujimoto, S., Hoof, H., and Meger, D.
\newblock Addressing function approximation error in actor-critic methods.
\newblock In \emph{International conference on machine learning}, pp.\
  1587--1596. PMLR, 2018.

\bibitem[Fujimoto et~al.(2019{\natexlab{a}})Fujimoto, Conti, Ghavamzadeh, and
  Pineau]{fujimoto2019benchmarking}
Fujimoto, S., Conti, E., Ghavamzadeh, M., and Pineau, J.
\newblock Benchmarking batch deep reinforcement learning algorithms.
\newblock \emph{arXiv preprint arXiv:1910.01708}, 2019{\natexlab{a}}.

\bibitem[Fujimoto et~al.(2019{\natexlab{b}})Fujimoto, Meger, and Precup]{BCQ}
Fujimoto, S., Meger, D., and Precup, D.
\newblock Off-policy deep reinforcement learning without exploration.
\newblock In \emph{International conference on machine learning}, pp.\
  2052--2062. PMLR, 2019{\natexlab{b}}.

\bibitem[Ghosh et~al.(2019)Ghosh, Gupta, Reddy, Fu, Devin, Eysenbach, and
  Levine]{GCSL}
Ghosh, D., Gupta, A., Reddy, A., Fu, J., Devin, C., Eysenbach, B., and Levine,
  S.
\newblock Learning to reach goals via iterated supervised learning.
\newblock \emph{arXiv preprint arXiv:1912.06088}, 2019.

\bibitem[Kostrikov et~al.(2021{\natexlab{a}})Kostrikov, Fergus, Tompson, and
  Nachum]{kostrikov2021offline}
Kostrikov, I., Fergus, R., Tompson, J., and Nachum, O.
\newblock Offline reinforcement learning with fisher divergence critic
  regularization.
\newblock In \emph{International Conference on Machine Learning}, pp.\
  5774--5783. PMLR, 2021{\natexlab{a}}.

\bibitem[Kostrikov et~al.(2021{\natexlab{b}})Kostrikov, Nair, and Levine]{IQL}
Kostrikov, I., Nair, A., and Levine, S.
\newblock Offline reinforcement learning with implicit q-learning.
\newblock \emph{arXiv preprint arXiv:2110.06169}, 2021{\natexlab{b}}.

\bibitem[Kumar et~al.(2019)Kumar, Fu, Soh, Tucker, and Levine]{BEAR}
Kumar, A., Fu, J., Soh, M., Tucker, G., and Levine, S.
\newblock Stabilizing off-policy q-learning via bootstrapping error reduction.
\newblock \emph{Advances in Neural Information Processing Systems}, 32, 2019.

\bibitem[Kumar et~al.(2020)Kumar, Zhou, Tucker, and Levine]{CQL}
Kumar, A., Zhou, A., Tucker, G., and Levine, S.
\newblock Conservative q-learning for offline reinforcement learning.
\newblock \emph{Advances in Neural Information Processing Systems},
  33:\penalty0 1179--1191, 2020.

\bibitem[LeCun et~al.(2015)LeCun, Bengio, and Hinton]{DL}
LeCun, Y., Bengio, Y., and Hinton, G.
\newblock Deep learning.
\newblock \emph{nature}, 521\penalty0 (7553):\penalty0 436--444, 2015.

\bibitem[Levine et~al.(2020)Levine, Kumar, Tucker, and Fu]{offline1}
Levine, S., Kumar, A., Tucker, G., and Fu, J.
\newblock Offline reinforcement learning: Tutorial, review, and perspectives on
  open problems.
\newblock \emph{arXiv preprint arXiv:2005.01643}, 2020.

\bibitem[Liu et~al.(2022{\natexlab{a}})Liu, Feng, Liu, and Stone]{MRN}
Liu, B., Feng, Y., Liu, Q., and Stone, P.
\newblock Metric residual networks for sample efficient goal-conditioned
  reinforcement learning.
\newblock \emph{arXiv preprint arXiv:2208.08133}, 2022{\natexlab{a}}.

\bibitem[Liu et~al.(2022{\natexlab{b}})Liu, Zhu, and Zhang]{GCRL}
Liu, M., Zhu, M., and Zhang, W.
\newblock Goal-conditioned reinforcement learning: Problems and solutions.
\newblock \emph{arXiv preprint arXiv:2201.08299}, 2022{\natexlab{b}}.

\bibitem[Ma et~al.(2021)Ma, Yang, Hu, Liu, Yang, Zhang, Zhao, and Liang]{VEM}
Ma, X., Yang, Y., Hu, H., Liu, Q., Yang, J., Zhang, C., Zhao, Q., and Liang, B.
\newblock Offline reinforcement learning with value-based episodic memory.
\newblock \emph{arXiv preprint arXiv:2110.09796}, 2021.

\bibitem[Ma et~al.(2022)Ma, Yan, Jayaraman, and Bastani]{GoFAR}
Ma, Y.~J., Yan, J., Jayaraman, D., and Bastani, O.
\newblock Offline goal-conditioned reinforcement learning via \$f\$-advantage
  regression.
\newblock In Oh, A.~H., Agarwal, A., Belgrave, D., and Cho, K. (eds.),
  \emph{Advances in Neural Information Processing Systems}, 2022.
\newblock URL \url{https://openreview.net/forum?id=_h29VprPHD}.

\bibitem[Mao et~al.(2022)Mao, Yang, Zhang, and Basar]{game4}
Mao, W., Yang, L., Zhang, K., and Basar, T.
\newblock On improving model-free algorithms for decentralized multi-agent
  reinforcement learning.
\newblock In \emph{International Conference on Machine Learning}, pp.\
  15007--15049. PMLR, 2022.

\bibitem[Mnih et~al.(2013)Mnih, Kavukcuoglu, Silver, Graves, Antonoglou,
  Wierstra, and Riedmiller]{mnih2013playing}
Mnih, V., Kavukcuoglu, K., Silver, D., Graves, A., Antonoglou, I., Wierstra,
  D., and Riedmiller, M.
\newblock Playing atari with deep reinforcement learning.
\newblock \emph{arXiv preprint arXiv:1312.5602}, 2013.

\bibitem[Nair et~al.(2020)Nair, Gupta, Dalal, and Levine]{AWAC}
Nair, A., Gupta, A., Dalal, M., and Levine, S.
\newblock Awac: Accelerating online reinforcement learning with offline
  datasets.
\newblock \emph{arXiv preprint arXiv:2006.09359}, 2020.

\bibitem[Nguyen \& La(2019)Nguyen and La]{robot2}
Nguyen, H. and La, H.
\newblock Review of deep reinforcement learning for robot manipulation.
\newblock In \emph{2019 Third IEEE International Conference on Robotic
  Computing (IRC)}, pp.\  590--595. IEEE, 2019.

\bibitem[Peng et~al.(2020)Peng, Kumar, Zhang, and Levine]{AWR}
Peng, X.~B., Kumar, A., Zhang, G., and Levine, S.
\newblock Advantage weighted regression: Simple and scalable off-policy
  reinforcement learning, 2020.
\newblock URL \url{https://openreview.net/forum?id=H1gdF34FvS}.

\bibitem[Peters \& Schaal(2007)Peters and Schaal]{RWR}
Peters, J. and Schaal, S.
\newblock Reinforcement learning by reward-weighted regression for operational
  space control.
\newblock In \emph{Proceedings of the 24th international conference on Machine
  learning}, pp.\  745--750, 2007.

\bibitem[Plappert et~al.(2018)Plappert, Andrychowicz, Ray, McGrew, Baker,
  Powell, Schneider, Tobin, Chociej, Welinder, et~al.]{Tasks}
Plappert, M., Andrychowicz, M., Ray, A., McGrew, B., Baker, B., Powell, G.,
  Schneider, J., Tobin, J., Chociej, M., Welinder, P., et~al.
\newblock Multi-goal reinforcement learning: Challenging robotics environments
  and request for research.
\newblock \emph{arXiv preprint arXiv:1802.09464}, 2018.

\bibitem[Prudencio et~al.(2022)Prudencio, Maximo, and
  Colombini]{prudencio2022survey}
Prudencio, R.~F., Maximo, M.~R., and Colombini, E.~L.
\newblock A survey on offline reinforcement learning: Taxonomy, review, and
  open problems.
\newblock \emph{arXiv preprint arXiv:2203.01387}, 2022.

\bibitem[Silver et~al.(2014)Silver, Lever, Heess, Degris, Wierstra, and
  Riedmiller]{dpg}
Silver, D., Lever, G., Heess, N., Degris, T., Wierstra, D., and Riedmiller, M.
\newblock Deterministic policy gradient algorithms.
\newblock In \emph{International conference on machine learning}, pp.\
  387--395. PMLR, 2014.

\bibitem[Sutton \& Barto(2018)Sutton and Barto]{RL}
Sutton, R.~S. and Barto, A.~G.
\newblock \emph{Reinforcement learning: An introduction}.
\newblock MIT press, 2018.

\bibitem[Wang et~al.(2020)Wang, Novikov, Zolna, Merel, Springenberg, Reed,
  Shahriari, Siegel, Gulcehre, Heess, et~al.]{CRR}
Wang, Z., Novikov, A., Zolna, K., Merel, J.~S., Springenberg, J.~T., Reed,
  S.~E., Shahriari, B., Siegel, N., Gulcehre, C., Heess, N., et~al.
\newblock Critic regularized regression.
\newblock \emph{Advances in Neural Information Processing Systems},
  33:\penalty0 7768--7778, 2020.

\bibitem[Wu et~al.(2019)Wu, Tucker, and Nachum]{wu2019behavior}
Wu, Y., Tucker, G., and Nachum, O.
\newblock Behavior regularized offline reinforcement learning.
\newblock \emph{arXiv preprint arXiv:1911.11361}, 2019.

\bibitem[Yang et~al.(2022)Yang, Lu, Li, Sun, Fang, Du, Li, Han, and
  Zhang]{WGCSL}
Yang, R., Lu, Y., Li, W., Sun, H., Fang, M., Du, Y., Li, X., Han, L., and
  Zhang, C.
\newblock Rethinking goal-conditioned supervised learning and its connection to
  offline rl.
\newblock \emph{arXiv preprint arXiv:2202.04478}, 2022.

\end{thebibliography}
\bibliographystyle{icml2023}

%%%%%%%%%%%%%%%%%%%%%%%%%%%%%%%%%%%%%%%%%%%%%%%%%%%%%%%%%%%%%%%%%%%%%%%%%%%%%%%
%%%%%%%%%%%%%%%%%%%%%%%%%%%%%%%%%%%%%%%%%%%%%%%%%%%%%%%%%%%%%%%%%%%%%%%%%%%%%%%
% APPENDIX
%%%%%%%%%%%%%%%%%%%%%%%%%%%%%%%%%%%%%%%%%%%%%%%%%%%%%%%%%%%%%%%%%%%%%%%%%%%%%%%
%%%%%%%%%%%%%%%%%%%%%%%%%%%%%%%%%%%%%%%%%%%%%%%%%%%%%%%%%%%%%%%%%%%%%%%%%%%%%%%
\newpage
\appendix
\onecolumn
\section{Algorithm Derivation Details}
\label{subsec:adv_obj_inference}

\subsection{The advantage optimization objective}
%Based on the policy gradient theorem, maximizing $Q(s,a)$ is equal to maximizing advantage function $A(s,a)$~\cite{RL}.
The most popular way to solve the distributional shift issue (extrapolation error) is  to add policy constraints to the policy improvement update. In this section, we present the details of the objective in Eq.\ref{eq:adv_bc} by strictly following work ~\cite{AWR}.  The optimization problem in goal-conditioned settings can be formulated as~\cite{AWR, AWAC}: 
\begin{equation}
\label{eq:a_orl_obj}
    \underset{\pi}{\argmax} = \int_g \rho_D(g) \int_s d_D(s|g) \int_a [A^{\pi}(s,g,a)]\textbf{\textit{d}}a\textbf{\textit{d}}s\textbf{\textit{d}}g,\quad \int_{a} \pi(a|s,g) \textbf{\textit{d}}a=1, \quad  \textbf{s.t.} \kl(\pi_\psi \lVert \pi_D) \leq \epsilon,
\end{equation}
where $\rho_D(g)$ represents the goal distribution in offline set $D$, and  
$d_D(s|g) = \sum_{t=0}^{\inf} \gamma^t p(s_t=s|\pi_D,g)$
represents the goal-conditioned unnormalized discounted state distribution induced by the offline dataset policy $\pi_D$~\cite{RL} , $\pi_\psi$ is the learned policy, $D$ is the offline goal-conditioned dataset, $\pi_D$ represents the behavioral policy that collects the dataset,  $\mathcal{D(\lVert)}$ is some divergence measurement metric, $\epsilon$ is the threshold. 
We can enforce the KKT condition on Eq.\ref{eq:a_orl_obj}   and get the Lagrangian:
\begin{equation}
\begin{split}
    \mathcal{L}(\lambda, \pi, \alpha) =  \int_g \rho_D(g) \int_s d_D(s|g)\int_a \pi(a|s,g)[A^{\pi_{}}(s,g,a)] \textbf{\textit{d}}a \textbf{\textit{d}}s \textbf{\textit{d}}g \\
    + \lambda(\epsilon - \int_g \rho_D(g) \int_s d_D(s|g) \kl(\pi_\psi \lVert \pi_D)\textbf{\textit{d}}s)\textbf{\textit{d}}g \\
    + \int_s \alpha(1- \int_{a} \pi(a|s,g) \textbf{\textit{d}}a) \textbf{\textit{d}}s
\end{split}
\end{equation}
Differentiating with respect to $\pi$ :
\[
\frac{\partial \mathcal{L}}{\partial \pi} = p_D(s,g) A^{\pi}(s,g,a) - \lambda \cdot p_D(s,g) \cdot \log \pi_D(a|s,g) + \lambda p_D(s,g) \log \pi(a|s,g) + \lambda p_D(s, g) - \alpha 
\]
where $p_D(s,g)=\rho_D(g)d_D(s|g)$. Then set $\frac{\partial \mathcal{L}}{\partial \pi}$ to zero we can have
\begin{equation}
\pi(a|s,g) = \pi_D(a|s,g) \exp (\frac{1}{\lambda} A^{\pi}) \exp (-\frac{1}{p_D (s,g)} \frac{\alpha}{\lambda} - 1).
\end{equation}
 $\exp (-\frac{1}{p_D (s,g)} \frac{\alpha}{\lambda} - 1)$ term is the partition function $Z(s,g)$ that normalizes a
the goal-state conditional action distribution~\cite{AWR}:
\begin{equation}
Z(s,g) = \exp (-\frac{1}{p_D (s,g)} \frac{\alpha}{\lambda} - 1) = \int_a' \pi_D(a'|s,g) \exp(\frac{1}{\lambda}A^\pi(s,g,a'))\textbf{\textit{d}}a'.
\end{equation}
The closed-form solution can then be represented as:
\begin{equation}
    \pi^*(a|s,g) = \frac{1}{Z(s,g)}\pi_D(a|s,g) \exp (\frac{1}{\lambda} A^\pi(s,g,a))
\end{equation}
Finally, as $\pi$ is a parameterized function, we project the solution into the space of parametric policies and get the following optimization objective:

 \begin{equation}
\begin{split}
    \underset{\pi}{\argmin} \mathbb{E}_{s,g \sim D} [\kl(\pi^*(\cdot|s,g) \lVert \pi(\cdot | s,g))] 
   &= \underset{\pi}{\argmin} \mathbb{E}_{s,g \sim D} [\kl(\frac{1}{Z(s,g)}\pi_D(a|s,g) \exp (\frac{1}{\lambda} A^\pi(s,g,a)) \lVert \pi(\cdot | s,g))] \\
    &= \underset{\pi}{\argmin} \mathbb{E}_{s,g \sim D}\mathbb{E}_{a \sim D} [- \pi(a|s,g) \exp (\frac{1}{\lambda} A^\pi(s,g,a)) ]
\end{split}
\end{equation}
Now this objective can be seen as weighted maximum likelihood. For a deterministic policy, the objective becomes:
\begin{equation}
    \underset{\pi}{\argmin} \mathbb{E}_{s,g, a \sim D} [ \exp (\frac{1}{\lambda} A^\pi(s,g,a) \cdot \MSE(\pi(s,g),a))
\end{equation}
However, $Z(s)$ is always disregarded in practical implementations ~\cite{AWR,AWAC}. Especially the work~\cite{AWAC} generally claims $Z(s,g)$ made performance worse. Thus we also ignored the term $Z(s)$ for practical implementation.

\subsection{The Q  optimization objective}
If we choose to optimize $Q(s,g,a)$ in Eq.~\ref{eq:policy_constraint_obj}, we have the policy improvement objective: 
\begin{equation}
\label{eq:orl_obj}
    \pi_{k+1} = \underset{\pi}{\argmax}\,\mathbb{E}[Q^{\pi_{k}}(s,g,\pi_k(s,g))], \quad \int_{a} \pi(a|s,g) \textbf{\textit{d}}a=1, \quad  \textbf{s.t.} \kl(\pi_\psi \lVert \pi_D) \leq \epsilon.
\end{equation}
As minimizing the KL-divergence is equal to minimizing the maximum likelihood~\cite{DL}, and considering a deterministic policy,  we have the following Lagrangian:
\begin{equation}
\begin{split}
    \mathcal{L}(\lambda, \pi) &=  \mathbb{E}_{a \sim \pi(\cdot|s)}[Q^{\pi}(s,g,a)] 
    + \lambda  \mathbb{E}_{a,s,g \sim D} \MLE(\pi(a|s,g)) %\\
    %&=  \mathbb{E}_{a \sim \pi(\cdot|s)}[Q^{\pi}(s,g,a)] - \lambda  \mathbb{E}_{a,s,g \sim D} \MSE(\pi(s,g),a)
\end{split}
\end{equation}

In this case, the deterministic policy $\pi(s,g)$ can be considered as a Dirac-Delta distribution thus, the $\int_{a}\pi(a|s,g)\textbf{\textit{d}}a=1$  constraint always satisfies.  The optimization objective is:
\begin{equation}
    \underset{\pi}{\argmin} \mathbb{E}_{s,g,a\sim D}[ -Q^\pi(s,g,a) + \MSE(\pi(s,g),a)]  
\end{equation}
Instead of finding a closed-form solution, this problem can be optimized through gradient descent. Based on the deterministic policy gradient theorem~\cite{dpg}, $Q$ function directly provides gradients for policy optimization.

\section{Implementation Details}

\label{subsec:exp_details}

\subsection{Offline RL technical details}
\paragraph{DQAPG}
As DQAPG is built upon the DPG framework, practically, we integrate our method into the TD3 framework~\cite{TD3}. In TD3, the double $Q$ networks and double $V$ networks can avoid the value overestimation issue. The detailed algorithm is presented in Algo.\ref{code:full-code} and Fig.\ref{fig:DQAPF-code}.

\paragraph{AM.} In actionable models' work, the policy is represented as:
\[
\pi(a|s,g)= \underset{a\sim D}{\argmax} Q^*(s,g,a),
\]
which only samples actions from the dataset and finds the action with maximum return. The objective of the value function is defined as:
\[\operatorname*{min}_{\theta}\mathbb{E}_{(s_{t},a_{t},s_{t+1},g)\sim{\mathcal D}}[((Q_{\theta}(s_{t},a_{t},g)-y(s_{t+1},g))^{2} + \mathbb{E}_{\tilde{a} \sim \mathrm{exp}(Q_{\theta})}[(Q_{\theta}(s,\tilde{a},g)-0)^{2}]]\]
where the TD target is:
\[
y(s_{t+1},g)=\left\{\begin{array}{l l}{{1}}&{{\mathrm{if~}s_{t+1}=g}}\\ {{\gamma \mathbb{E}_{a\sim\pi}[Q(s_{t+1},a,g)]}}&{{\mathrm{otherwise.}}}\end{array}\right.
\]
The first part of the loss $(Q_{\theta}(s_{t},a_{t},g)-y(s_{t+1},g))^{2}$ increases Q-values for reachable goals and the second part $\mathbb{E}_{\tilde{a} \sim \mathrm{exp}(Q_{\theta})}[(Q_{\theta}(s,\tilde{a},g)-0)^{2}$ regularizes the Q-function. When optimizing with gradient descent, the loss does not propagate through the sampling of action negatives $\tilde{a}\sim \exp$.
We use the AM model that is implemented work~\cite{GoFAR}.  10 random actions are sampled from the action-space to approximate this expectation. We keep the goal-chaining (goal-swapping) for all experiments. The hyperparameters are presented in Sec.\ref{subsec:Hype}.

\paragraph{GoFAR, GCSL and WGCSL.} GCSL is a simple supervised regression method that does behavior cloning from hindsight relabelling dataset. Thus we only use the policy network from TD3 and optimize it with maximum likelihood loss. We refer the objective to Table \ref{tab:algo_comp}. 
The WGCSL is implemented on top of GCSL, incorporated with TD3. More specifically, the policy objective is:
\[\min_{\pi} \mathbb{E}_{(s_t,a_t,g)\sim D_{relabel}}[-\gamma^{t-i}A_{wgcsl}^\pi(s_t,g,a_t) \cdot \log \pi(a_t|s_t, g)],\]
where $i$ represents the hindsight relabelling steps in the future. The advantage weighting in the regression loss is the TD error
\[
A_{wgcsl}^\pi(s,g,a) = r+Q^\pi(s',g,\pi(s',g)) - Q(s,g,a), \quad a\sim D. 
\] As AM, GoFAR, WGCSL and GCSL are fine-tuned in work~\cite{GoFAR}, we keep the same implementations for this work \url{https://github.com/jasonma2016/gofar}.

%\paragraph{GoFAR} As GoFAR optimizes the policy through state-occupancy measurement, it requires an additional discriminator for training.

\paragraph{BCQ}
The goal-conditioned objective function of  BCQ in this work is as follows:
\[
\pi(a|s,g)= \underset{{a\sim[\vae(s,g)+\sigma_\phi(s,g,a)]}}{\argmax}  Q^*(s,g,a) 
\]
where $\sigma_\phi(s,g,a)$ is the goal conditioned perturbation model, which disturbs the action within a range between $[-\Phi, \Phi]$. The generative model $\vae(s,g)$ generates the actions that fit the offline dataset's distribution. The Q function is learned using the Clipped Double Q-learning in TD3~\cite{TD3}. The policy is implicitly presented as the above equation: sample the actions within the dataset with the maximum return. All hyperparameters are kept the same as the original work~\cite{BCQ}. The only difference is that we have an additional goal as input. In this work, we implemented BCQ based on its official source code \url{https://github.com/sfujim/BCQ}.

\paragraph{CQL} 
In work CQL, the constrain $\kl(\pi(\cdot|s,g) \lVert \pi_D(\cdot|s,g))$ is added on value function instead of directly on policy. The loss of the value function can be written as:
\begin{equation}
    \begin{split}
        L_{CQL}
        =\alpha\mathbb{E}_{(s_{t},g)\sim D}\big[\log\sum_{a}\exp(Q(s_{t},a,g))-\mathbb{E}_{a\sim\pi_{b}(\cdot|s_{t},g)}|Q(s_{t},a,g)\big])\\  
        +\;\frac{1}{2}\mathbb{E}_{(s_{t},a_{t},s_{t+1},g)\sim D}\left[(Q(s_{t},a_{t},g)-\mathcal{B}^{\pi}Q(s_{t},a_{t},g))^{2}\right],
    \end{split}
\end{equation}
where $\pi_b$ is the behavior policy that collects offline dataset $D$ and $\mathcal{B}^{\pi}$ is the Bellman operator. The policy objective of CQL is a typical policy gradient objective, which is formulated as:
\[\min_{\pi}\mathbb{E}_{{s,g}\sim D, a~\sim\pi(\cdot|s,g)} [- \log \pi_\psi(a|s,g) Q^\pi(s,g,a)] \]
All hyperparameters are kept the same as the original work~\cite{BCQ}. The only difference is that we have an additional goal as input.  The technical implementation is based on the official code \url{https://github.com/aviralkumar2907/CQL}.

\paragraph{CRR} The idea of CRR is to use $Q(s,a,g)$ to weight the behavior cloning, and the policy objective is :
\[
\underset{\pi}{\argmax} \mathbb{E}_{s,a,g \sim D} [f( Q^\pi_{\theta},\pi,s,a)\,\mathrm{log}\,\pi(a|s,g)],
\]
where $f$ is a non-negative, scalar, monotonically increasing function in $Q_\theta$. In this work, we choose to use:
\[
f = \mathbb{I}[A^\pi(s,g,a)>0]
\] as it has the best performance in the original implementation. $\mathbb{I}$ is the indicator function. Our CRR is implemented based on \url{https://github.com/takuseno/d3rlpy}, and all hyperparameters are kept the same as the original work~\cite{CRR}.

\paragraph{AWAC} The work AWAC shares a similar idea as CQL, but it uses the advantage function to weight the regression. Its policy objective is:
\[
\underset{\pi}{\argmax} \mathbb{E}_{s,a,g \sim D} [\exp( A_{\theta}^\pi(s,g,a)\,\mathrm{log}\,\pi(a|s,g)],
\] and the advantage value is the TD error: 
\[
A^\pi_\theta(s,g,a) = r+Q^\pi_\theta(s',g,a') - Q^\pi_\theta(s,g,a), \quad a\sim D, a' \sim \pi(\cdot|s,g). 
\].  Our CRR is implemented based on \url{https://github.com/takuseno/d3rlpy}.

\paragraph{Onsetep-RL, IQL} The idea of Onestep-RL is only learning $Q^{\pi_{D}}$ instead of $Q^*$. It means that the algorithm only conducts  policy evaluation based on offline dataset $D$ once, and use the estimated value functions to optimize $\pi$ ~\cite{onestep}. In such case, we can write the policy objective as:
\[
\underset{\pi}{\argmax} \mathbb{E}_{s,g \sim D, a\sim \pi(\cdot|s,g)} [\exp( A_{\theta}^{\textcolor{red}{{\pi_D}}}(s,g,a)\,\mathrm{log}\,\pi(a|s,g)],
\] where $\textcolor{red}{{\pi_D}}$ is the behavior policy that collects dataset $D$. To improve the performance of Onestep-RL, the work IQL proposes using an expectile regression loss for policy evaluation,  which is defined as:
\begin{equation*}
       \mathcal{L}^\tau_2(x)= 
\begin{cases}
    (1-\tau)x^2, & \text{if } x>0\\
    \tau x^2,              & \text{otherwise}
\end{cases},
\end{equation*} 
where $\tau$ represents the expectile. In this way, IQL considers $V^\tau (s,g), Q^\tau (s,g,a), A^\tau(s,g,a)$ as the best values from the actions within the support of offline data ~\cite{IQL}. The IQL has the same policy objective as Onestep-RL, and its value function objective can be written as:
\[
{\cal L}_{V} (\omega)= \mathbb{E}_{(s,g,a)\sim D} [\mathcal{L}_2^\tau(Q_\theta(s,a)-V_\omega(s))]
\]
\[
{\cal L}_{Q}= \mathbb{E}_{(s,g,a,s')\sim D} [(r(s,g,a) + \gamma V_\omega(s,g) - Q_\theta(s,g,a) )^2]
\]

Note that when the $\tau=0.5$, IQL is identical to Onestep RL.  In this work, we choose $\tau=0.75$ as it has the best performance~\cite{IQL}. We use the official implementation in this work ~\url{https://github.com/ikostrikov/implicit_q_learning}.

\subsection{Hyperparameters}
\label{subsec:Hype}
In this work, DQAPG, GCSL, WGCSL, GoFAR, CRR, AWAC, Onestep-RL, IQL, AM, TD3BC share the same network architectures. The hyperparameters of these methods are presented in Table.\ref{tab:hype}. In work~\cite{GoFAR}, GoFAR, WGCSL, GCSL, and AM's parameters have already been fine-tuned for goal-conditioned tasks. Thus we keep the hyperparameters as implemented in ~\cite{GoFAR}. For BCQ and CQL implementation, we keep the same hyperparameters used in their original works ~\cite{BCQ, CQL}.

\begin{table*}[t]
\caption{General hyperparameters of networks 
} 
\label{tab:hype}

\begin{center}
\resizebox{0.35\columnwidth}{!}{
  \begin{tabular}{lcc}
    \toprule
    \textbf{Hyperparameter} & \textbf{Value}  \\
     % & Imitation return & True return  & Imitation return & True return & Imitation return & True return & Imitation return & True return & Imitation return & True return\\
    \midrule
%Surface-follow & -1.23  & - &  -  & -  \\
%\midrule

Number of layers ($\pi$, $Q$, $V$ ) & 3\\
Size of hidden layers ($\pi$, $Q$, $V$) & 256\\
Activation functions ($\pi$, $Q$, $V$) & ReLU \\
Batch size & 512 \\
Learning rate & $10^{-3}$\\
Polyak average weight & 0.95\\
Target network update steps & 10\\
Hindsight relabelling ratio & 0.5\\

    \bottomrule
  \end{tabular}}
\end{center}
\end{table*}

\section{Task details}

\subsection{Fetch Tasks}
For Fetch tasks, the environments are based on the 7-DoF Fetch robotics arm, which has a two-fingered parallel gripper. The robot action frequency is set as $25 Hz$ ~\cite{Tasks}. The state includes 1) the object's Cartesian position and rotation using Euler angles, 2) its linear and angular velocities, and 3) its position and linear velocities relative to the gripper. A goal is considered achieved if the distance between the block's position and its desired position is
less than 1 cm. The environment's original binary reward is defined as
\begin{equation}
        r= 
\begin{cases}
    1, & \text{if } distance\,\,< 1cm\\
    0,              & \text{otherwise}
\end{cases}.
\end{equation}
The environment's original binary reward is used for task performance evaluation (Section ~\ref{subsec:experiment}).
\begin{itemize}
    \item \textbf{FetchPickAndPlace} In this task, the task is to grasp the object and move it to the target location. The target location is randomly sampled in the space (e,g., on the table or above it). The goal is 3-dimensional and describes the desired position of the object.

    \item \textbf{FetchPush} In this task, a block is placed in front of the robot. The robot can only push or roll the object to the target position. The goal is 3-dimensional and describes the desired position of the object.
    
    \item \textbf{FetchSlide} The task is to move a block to a target position.  The block is put on a long slippery table, and the target position cannot be reached by the robot. The robot must hit the block so it can slide to the target position. The goal is 3-dimensional and describes the desired position of the object.
    
    \item \textbf{FetchReach} In this task, the robot is required to move its end-effector to a specific position. The goal is 3-dimensional and describes the desired position of the end-effector.
\end{itemize}

\subsection{In-hand manipulation tasks}
For in-hand manipulation tasks, the environments are based on s an anthropomorphic robotic hand with 24 degrees of freedom. The observation includes 24 positions and velocities of the robot's joints, a quaternion representation of the manipulated object's linear and angular velocities. The actions are implemented as absolute position control, which is 20-dimensional. The action frequency is set to $25 \mathrm{Hz}$. In these tasks,  an object is placed on the palm of the hand, and the task is to manipulate the block to reach the target pose. The goal is 7-dimensional and includes the target position (in Cartesian coordinates) and target rotation (in quaternions). No target position is given in all tasks.
A goal is considered achieved if the distance between the object’s rotation and its desired rotation is less than 0.1 rad. The environment's original binary reward is defined as
\begin{equation}
        r(s,g)= 
\begin{cases}
    1, & \text{if } rotation\,\, difference\,\, <\,\, 1 rad\\
    0,              & \text{otherwise}
\end{cases}.
\end{equation}
The environment's original binary reward is used for task performance evaluation (Section ~\ref{subsec:experiment}).
\begin{itemize}
    \item \textbf{HandManipulationBlockRotateZ} The task is to manipulate the block to a random target rotation around the z-axis of the block. 
    \item \textbf{HandManipulationBlockParallel} The task is to manipulate the block to a random target rotation around the z-axis of the block and axis-aligned target rotations for the x and y axes. 
    \item \textbf{HandManipulationBlock} The task is to manipulate the block to a random target rotation for all axes of the block. 
    \item \textbf{HandManipulationEggRotate} The task is to manipulate the egg-like object to a random target rotation for all axes of the egg.
\end{itemize}

\subsection{PointMaze}
\label{subsec:pointmaze_fig}
\begin{figure*}[h]
    \centering
    \includegraphics[width=.5\textwidth]{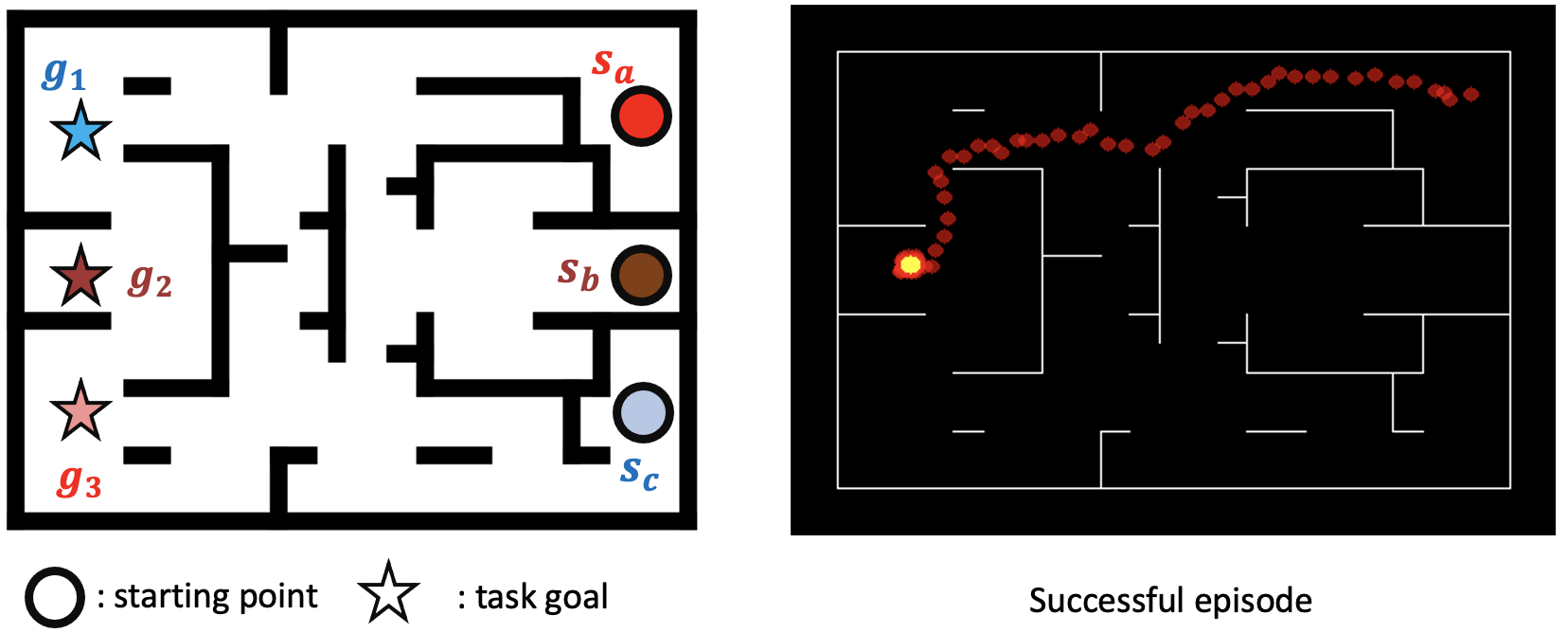}
    \caption{PointMaze environment. The left figure illustrates the task settings. There are 3 starting points and 3 task goals. The right figure presents one successful trajectory generated by an expert.}%$\tau^a$ and $\tau^b$ have mutual visited state $s_t$. The

    \label{fig:sup-point-maze}
\end{figure*}

We further implemented a customized continuous 2D PointMaze. In this environment, the agent is randomly placed at a starting point $s_0 \sim [s_a, s_b, s_c]$, and the task goal is randomly chosen from goal clusters $g \sim [g_1, g_2, g_3]$. Specifically, each starting point $s \in [s_a, s_b, s_c]$ shown in Figure~\ref{fig:sup-point-maze} represents a cluster of points, and so do the goal points. The maze size is $24 \times 18$. A goal is considered achieved if the euclidean distance is smaller than 2. The observation is the agent's position in this maze $(x,y)$, and the action is in the range $(2,2)$. We use this task to visualize the generalizability problem faced by goal-conditioned reinforcement learning.

\begin{figure*}[h]
    \centering
    \includegraphics[width=.95\textwidth]{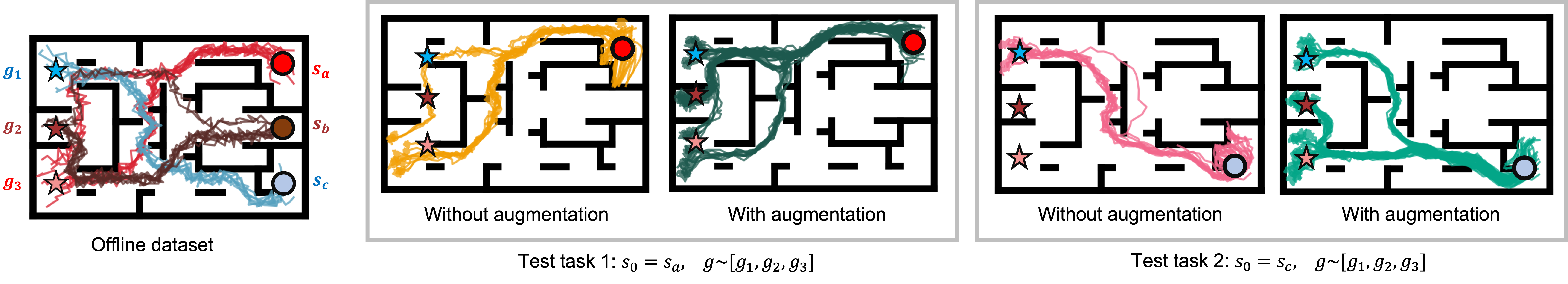}
    \caption{The PointMaze visually presents the generalizability issue of offline goal-conditioned RL. Without goal-swapping augmentation, the offline RL only learns an overfitted policy. }%$\tau^a$ and $\tau^b$ have mutual visited state $s_t$. The

    \label{fig:addition-maze}
\end{figure*}

\section{Dataset collection}

\label{subsec:dataset}
 \paragraph{PointMaze.\quad} We used $\mathrm{A}^*$ as an expert to generate the offline dataset for GCRL. We add random action noise on the expert-planned trajectories to generate sub-optimal training data. As is discussed in Sec.\ref{subsec:experiment}, only three types of goal-conditioned trajectories are collected, in total 30 trajectories.
 
\paragraph{Fetch tasks.\quad} For the Fetch tasks, we use the dataset collected in work WGCSL~\cite{WGCSL}, which is generally used as goal-conditioned offline RL benchmarks~\cite{GoFAR}. For each task, the original dataset contains 40K expert trajectories and 40K trajectories collected by random policy. As we address the data-efficiency and generalizability problem, in this work we only take 500 expert trajectories and 2K random trajectories to form a new offline GCRL dataset. The original dataset is provided by WGCSL in \url{ https://github.com/YangRui2015/AWGCSL}. 

\paragraph{In-hand manipulation tasks.\quad} In-hand manipulation tasks are challenging tasks that no dataset was provided in the previous offline GCRL works. We use the MRN method provided in ~\cite{MRN} to collect the expert trajectories for all in-hand manipulation tasks. 500 expert trajectories and 2K random trajectories are collected for each task. More specifically, the implementation code of the expert policy can be found here \url{https://github.com/Cranial-XIX/metric-residual-network}.

\section{Source code}
\begin{figure*}[h]
    \centering
    \includegraphics[width=0.95\textwidth]{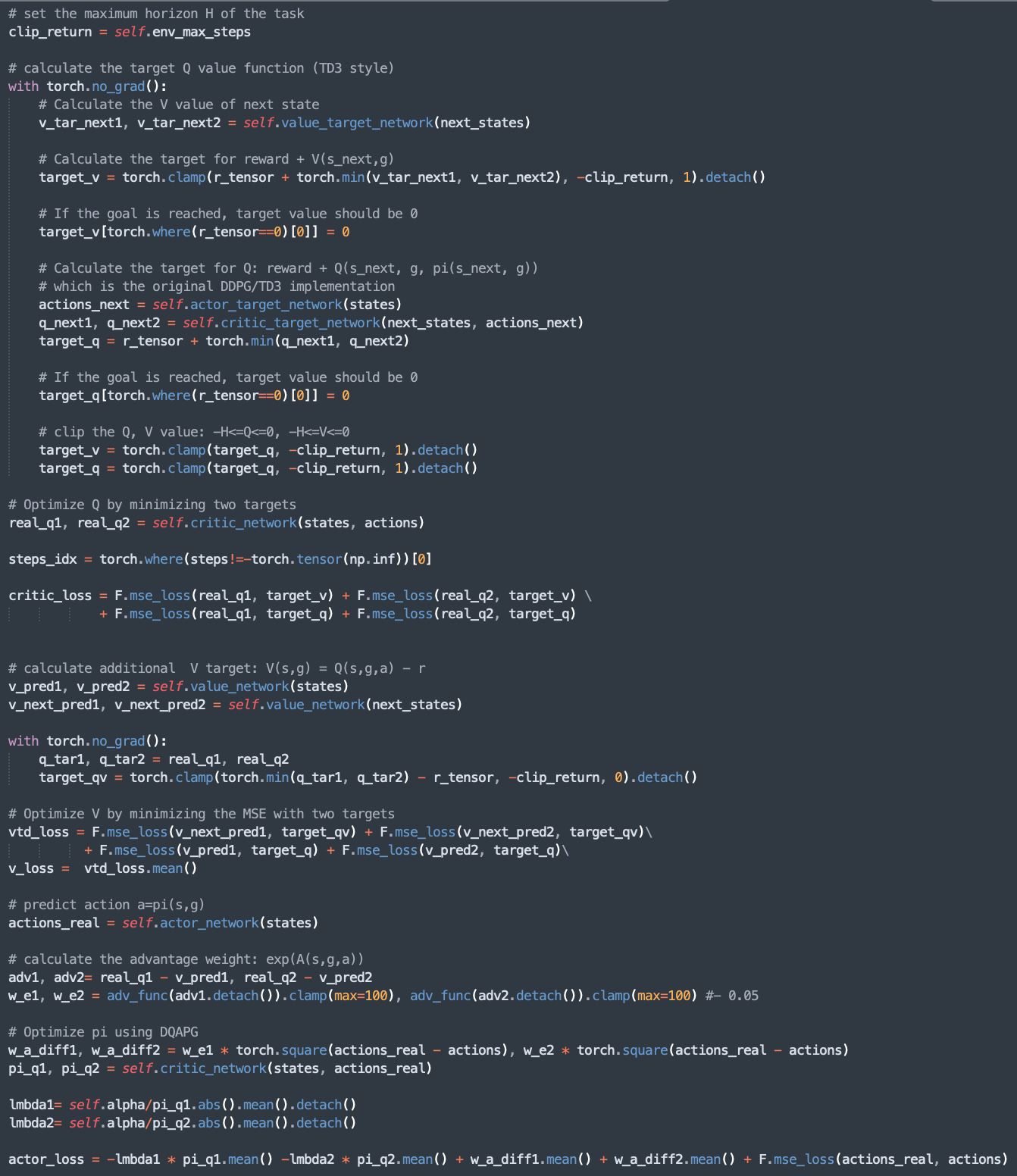}
    \caption{Core code of DQAPG.}%$\tau^a$ and $\tau^b$ have mutual visited state $s_t$. The

    \label{fig:DQAPF-code}
\end{figure*}

First, we illustrate the full pseudo-code of our algorithms. We also attach our source code of DQAPG in this section. We use Pytorch 1.13.1 for implementation. The core code is presented in Figure~\ref{fig:DQAPF-code}, and the full pseudo-code is presented in Algo.\ref{alg:DQAPG-full}.

\begin{algorithm}
\caption{Offline goal-conditioned Deterministic Q-Advantage policy gradient}\label{alg:DQAPG-full}
Given offline dataset $D$,
 Initialize policy $\pi_\psi$, Q-functions $\{Q_{\theta1}, Q_{\theta2}\}$, target Q-functions $\{Q_{\theta1}^{tar}, Q_{\theta2}^{tar}\}$,  V-functions $\{V_{\omega_1}, V_{\omega_2}\}$, target V-functions $\{V_{\omega_1}^{tar}, V_{\omega_2}^{tar}\}$. Denote the task's maximum horizon as $H$, reward function as $R(\phi(s), g)$, and $\phi(s)$ maps state $s$ to goal $g$, target network update step as $h$;
 \label{code:full-code}
 \begin{algorithmic}[1]
        
        \FOR{i in range(epochs)}
        \STATE \textcolor{blue}{// Goal-swapping augmented experience replay}
        %Denote the task goal as $g$, state as $s$, state's corresponding achieved goal as $ag$, action as $a$. Given a binary reward function $R(ag, g)$. 
        %\STATE Partially hindsight relabelling trajectories.
        \STATE Sample goal-conditioned transitions $\zeta_i = \{g,s_t,a_t,r_t,s_{t+1}\} \sim D$:

        \STATE Hindsight relabel half of the batch using future goals:
        $\zeta_i = \{(\phi(s_i),s_t,a_t,R(\phi(s_i), s_t),s_{t+1}),t\leq i \leq H\} \sim D$.

        \STATE Sample random goals: $g_{rand} \sim D$.
        \STATE Generate $\tau_{rand}$  by replacing $g$ with $g_{rand}$ in $\zeta$:
        $\zeta_{aug}= \{(g_{rand},s_t,a_t,R(g_{rand}, s_t),s_{t+1})\} \sim D$.
        \STATE Return $\zeta$ and $\zeta_{aug}$ for DQAPG optimization. 
        \STATE \quad
        \STATE \textcolor{blue}{// Value function optimization}
        \STATE Compute $Q$ targets:
        \[y_q(s,g,a) = r + (1-d) \underset{i=1,2}{\min} V_{\omega_i}^{tar}(s',g)  \]
        \STATE Optimize $Q$ by minimizing $\MSE$ loss:
        \[\underset{\theta_i, i = 1,2}{\argmin}\MSE(Q_{\theta_i}(s,g,a), y_q)\]
        \STATE Compute $V$ targets:
        \[y_v(s,g,a) = r + (1-d) \underset{i=1,2}{\min} Q_{\theta_i}^{tar}(s',g, \pi_\psi(s',g))  \]
        \STATE Optimize $V$ by minimizing $\MSE$ loss:
        \[\underset{\omega_i, i = 1,2}{\argmin}\MSE(V_{\omega_i}(s,g,a), y_v)\]

        \STATE \textcolor{blue}{// Policy optimization}
        \STATE Calculate the advantage weight:
        \[w_A = \exp (A(s,g,a)) = \exp (Q_{\theta_1}(s,g,a) - V_{\omega_1}(s,g))\]
        \STATE Clip the advantage weights:
        \[w_A = \min (w_A, 100)\]
        \STATE Policy optimization by deterministic policy gradient and advantage weighted regression:
        \[\underset{\psi}{\argmin}(-\lambda Q_{\theta1}(s,g,\psi(s,g)) + w_e \cdot \MSE(\pi_\psi(s,g), a)), \quad \lambda = \frac{1}{\frac{1}{N}\sum_{s_i,g,a_i}|Q(s_i, g, a_i)|}\]

        \STATE \textcolor{blue}{// Update the delayed target networks}
        \STATE if $i$ mod $h=0$:
         \[\omega_i^{tar} \leftarrow \rho \omega_i^{tar} + (1-\rho)\omega_i, \quad i=1,2\]
         \[\theta_i^{tar} \leftarrow \rho \theta_i^{tar} + (1-\rho)\theta_i, \quad i=1,2\]
         ($\rho$ here is the polyak averaging weight).

    \ENDFOR
    %\Until{Training is accomplished}
    \end{algorithmic} 
\end{algorithm}

%%%%%%%%%%%%%%%%%%%%%%%%%%%%%%%%%%%%%%%%%%%%%%%%%%%%%%%%%%%%%%%%%%%%%%%%%%%%%%%
%%%%%%%%%%%%%%%%%%%%%%%%%%%%%%%%%%%%%%%%%%%%%%%%%%%%%%%%%%%%%%%%%%%%%%%%%%%%%%%

%%%%%%%%%%%%%%%%%%%%%%%%%%%%%%%%%%%%%%%%%%%%%%%%%%%%%%%%%%%%%%%%%%%%%%%%%%%%%%%
%%%%%%%%%%%%%%%%%%%%%%%%%%%%%%%%%%%%%%%%%%%%%%%%%%%%%%%%%%%%%%%%%%%%%%%%%%%%%%%

\end{document}